# Predictors and Orchestrators: Parsimonious Machine Learning within an Agentic AI Harness for Multi-Horizon Karst Aquifer Forecasting

Pramod Lekhak[a], Chetan Sharma[b], Hakan Başağaoğlu[c], F. Paul Bertetti[c], Debaditya Chakraborty[a,*]

[a]*School of Civil and Environmental Engineering & Construction Management, University of Texas at San Antonio, San Antonio, TX 78249, USA*

[b]*Civil Engineering Department, Jodhpur Institute of Engineering and Technology, Jodhpur-342802, Rajasthan, India*

[c]*Edwards Aquifer Authority, 900 E. Quincy St., San Antonio, TX 78215, USA*

---

*Corresponding author.

*Email addresses:* Pramod.Lekhak@my.utsa.edu (Pramod Lekhak), chetan.sharma@jietjodhpur.ac.in (Chetan Sharma), hbasagaoglu@edwardsaquifer.org (Hakan Başağaoğlu), pbertetti@edwardsaquifer.org (F. Paul Bertetti), debaditya.chakraborty@utsa.edu (Debaditya Chakraborty)

**Highlights**

- Tree ensembles outperform deep learning across 1-12 week horizons
- Forecast-origin integrity prevents future-information leakage
- Threshold skill links hydrologic forecasts directly to drought action
- Agents automate operations while numerical decisions remain auditable
- Prospective verification closes the loop between forecasts and observations

**Abstract**

Forecasting karst aquifer dynamics is difficult because recharge responses are nonlinear, event-driven, and governed by strongly heterogeneous flow paths. This study develops and evaluates a deployment-aware framework for 1-12-week-ahead prediction of spring discharge and groundwater level using approximately 79 years of hydroclimatic observations from the Edwards Aquifer, Texas. Five model families were compared under a common temporal evaluation design: extreme gradient boosting, extremely randomized trees, long short-term memory, convolutional neural networks, and Transformers. Predictions were evaluated using coefficient of determination, Kling-Gupta efficiency, root-mean-square error, and agreement with operational drought thresholds. Extreme gradient boosting was consistently most reliable, with $R^2$ at least 0.97, 0.96, and 0.94 across 1-4-, 5-8-, and 9-12-week horizons, respectively, and greater than 90% critical-stage agreement at the first three drought stages across all horizons. Deep models were competitive at short horizons but degraded progressively and exhibited isolated failures at longer lead times. We attribute this contrast to an alignment between tree partitioning and low-dimensional, axis-aligned hydroclimatic predictors, together with the tendency of neural models to smooth irregular extremes. The validated models were embedded in a five-agent operational architecture that automates data acquisition, model assignment, deterministic prediction, threshold monitoring, prospective verification, literature retrieval, and reporting. The contribution is therefore a transferable framework joining parsimonious model selection, leakage-aware multi-horizon evaluation, decision-relevant threshold skill, and auditable agentic automation.

## 1. Introduction

**From hydrologic complexity to operational prediction:** Karst aquifers pose a distinctive forecasting challenge. Long periods of gradual recession can be interrupted by abrupt recharge events, while operational decisions often depend on whether flow or groundwater level crosses a regulatory threshold rather than on average predictive accuracy alone[1–4]. These characteristics create a demanding test for machine-learning models: a useful forecast must reproduce both the broad evolution of the aquifer and the short-lived extremes that determine drought-management actions. This study addresses that challenge in the Edwards Aquifer of south-central Texas. The aquifer supplies water to more than two million people while sustaining ecologically important spring systems. Its heterogeneous conduits, rapid recharge pathways, and nonlinear response to precipitation make it scientifically complex and operationally consequential.

Against this background, we ask a deliberately practical question: are computationally intensive sequence models necessary for forecasting karst-aquifer conditions 1–12 weeks ahead, or can parsimonious tree ensembles provide more accurate, stable, and operationally defensible predictions when the available inputs are compact, structured hydroclimatic variables?

**Complementary indicators of aquifer condition:** We examine two hydrologic variables that represent different but complementary expressions of aquifer state. Comal Springs discharge reflects the integrated surface-water response of the aquifer, whereas groundwater elevation at the J-17 index well represents artesian pressure and storage within the San Antonio Pool. Together, these indicators connect regional aquifer dynamics to two major management concerns: sustaining spring flow and maintaining groundwater availability.

Both variables are represented by weekly minimum. This aggregation is intentionally conservative. Weekly averages can conceal brief excursions below regulatory thresholds, whereas weekly minimums retain the low-flow and low-level conditions most relevant to drought restrictions.

The analysis uses approximately 79 years of hydroclimatic observations spanning 1946–2025, including precipitation, maximum and minimum air temperature, spring discharge, and groundwater elevation. The record is divided chronologically into training, validation, and testing periods. This temporal separation preserves the direction of time and reduces the risk that future information influences retrospective performance estimates.

**A common test across model paradigms and forecast horizons:** Five model families are evaluated under the same temporal protocol. XGBoost[5] and Extremely Randomized Trees[6] represent ensemble tree methods, while long short-term memory networks[7], convolutional neural networks[8], and Transformers[9] represent deep sequence-learning approaches.

Each model produces forecasts at 12 independent lead times, grouped into short-range forecasts of 1–4 weeks, medium-range forecasts of 5–8 weeks, and long-range forecasts of 9–12 weeks. All experiments follow forecast-origin integrity: a prediction may use only information that would have been available at or before the date on which the forecast was issued. This constraint is essential because apparently strong retrospective results can otherwise depend on meteorological or hydrologic information unavailable during real operation.

**The simpler model proves the more reliable one:** The results reveal a consistent model hierarchy. XGBoost provides the strongest and most stable performance across both hydrologic indicators and all three forecast ranges. Its coefficient of determination remains at least 0.97 over weeks 1–4, 0.96 over weeks 5–8, and 0.94 over weeks 9–12. It also exceeds 90% agreement with the first three operational critical stages, which trigger mandatory groundwater-pumping reductions.

The deep-learning models are competitive at some short lead times, but their performance becomes less reliable as the forecast horizon increases. More importantly, their errors are not limited to a gradual loss of average accuracy. The neural predictions increasingly smooth the hydrograph, attenuating high-flow peaks and elevating critically low conditions toward the historical mean. Isolated numerical failures also occur, including CNN collapse at week 7 for Comal Springs and week 9 for J-17.

This contrast is important because it challenges the assumption that more complex temporal architectures necessarily provide better environmental forecasts. In this application, model complexity is not a reliable proxy for forecast skill.

**From retrospective forecasting to an operational scientific system:** *Another challenge arises after a model has been evaluated: most forecasting studies end with a retrospective benchmark. The resulting model remains disconnected from the data feeds, monitoring procedures, verification records, and reporting workflows required for sustained use.*

*To address this gap, the validated models are embedded in an agent-coordinated operational prototype. Five specialized agents organize the recurring workflow:*

1. A data-ingestion agent retrieves U.S. Geological Survey (USGS) spring-discharge data, Edwards Aquifer Authority (EAA) groundwater observations, and National Oceanic and Atmospheric Administration (NOAA) meteorological information.
2. A model-orchestration agent maps validated models to the 24 combinations of site and forecast horizon.
3. A prediction-and-monitoring agent executes forecasts, evaluates critical-stage conditions, and manages verification records.
4. A literature-search agent retrieves scientific and operational information relevant to current aquifer and drought conditions.
5. A report-generation agent combines observations, predictions, performance statistics, threshold status, and supporting literature into timestamped reports.

An orchestrator coordinates these components, but it does not determine the numerical forecast. This distinction establishes an explicit scientific trust boundary. Language-model-based agents may select tools, coordinate workflows, synthesize information, and generate explanatory text. Data preprocessing, model assignment rules, inference, performance calculations, threshold classification, and forecast verification remain deterministic Python operations. Consequently, the same numerical inputs and model artifacts produce the same numerical outputs, independent of the language model used for orchestration.

**Turning deployment into a continuing experiment:** The operational framework also introduces prospective verification. Every forecast is archived at issuance, before its target observation is available. When the observation matures, it is matched to the immutable forecast record and used to update estimates of bias, magnitude error, and threshold agreement.

This closed-loop design addresses a limitation of retrospective testing. Historical experiments cannot fully reproduce changing data services, missing observations, stale caches, forecast-weather uncertainty, or shifts in environmental conditions. Prospective verification measures performance under those real operational constraints. The current implementation incorporates a live seven-day National Weather Service forecast into the week-one prediction; longer horizons rely on the meteorological assumptions supported by the deployed feature schema. These assumptions are recorded explicitly so that operational performance is not confused with an idealized hindcast.

**Scientific contribution:** The principal novelty of this work (outlined in Table 1) is therefore the integration of four elements that are often studied separately:

- a cross-paradigm test of tree ensembles and deep sequence models over 12 forecast horizons;
- a mechanistic explanation linking model performance to karst discontinuities and feature-space geometry;
- evaluation against both continuous metrics and management-relevant drought thresholds; and
- an auditable agent-coordinated workflow that preserves deterministic numerical control and prospectively verifies issued forecasts.

Together, these elements shift the emphasis from identifying the most sophisticated model to constructing the most trustworthy forecasting process. For operational groundwater management, that distinction is fundamental: a useful system must not only predict well in a retrospective experiment, but also preserve forecast provenance, expose its assumptions, recognize decision thresholds, and learn from its performance after deployment.

Table 1: Principal scientific and methodological contributions.

| Contribution | Advance beyond a conventional case study |
|---|---|
| Cross-paradigm, multi-horizon benchmark | Five model families, two hydrologic targets, and twelve independently evaluated lead times under one temporal design. |
| Forecast-origin integrity | All retrospective predictors are restricted to information available at or before the forecast origin, preventing future-data leakage. |
| Model-data alignment | Performance differences are interpreted through the geometry and event structure of low-dimensional karst hydroclimatic data. |
| Decision-relevant verification | Continuous-value metrics are complemented by agreement with operational drought thresholds. |
| Auditable agentic operations | Agents coordinate ingestion and reporting, while feature construction, model selection, inference, thresholds, and verification remain deterministic. |
| Prospective evidence loop | Issued predictions are archived before outcomes occur and scored when observations become available. |

## 2. Study area and data

The Edwards Aquifer in south-central Texas (Figure 1) is one of the largest and most productive karst aquifers globally. Composed of heterogeneous, faulted, and fractured limestone, it exhibits nonlinear groundwater flow through multiscale pathways ranging from microchannels to large conduits[10]. Extending from a semi-arid climate in the west to a sub-humid climate in the east, the aquifer supplies water to more than 2 million people and sustains key spring ecosystems, including Comal and San Marcos Springs, which serve as critical habitats for threatened and endangered species.

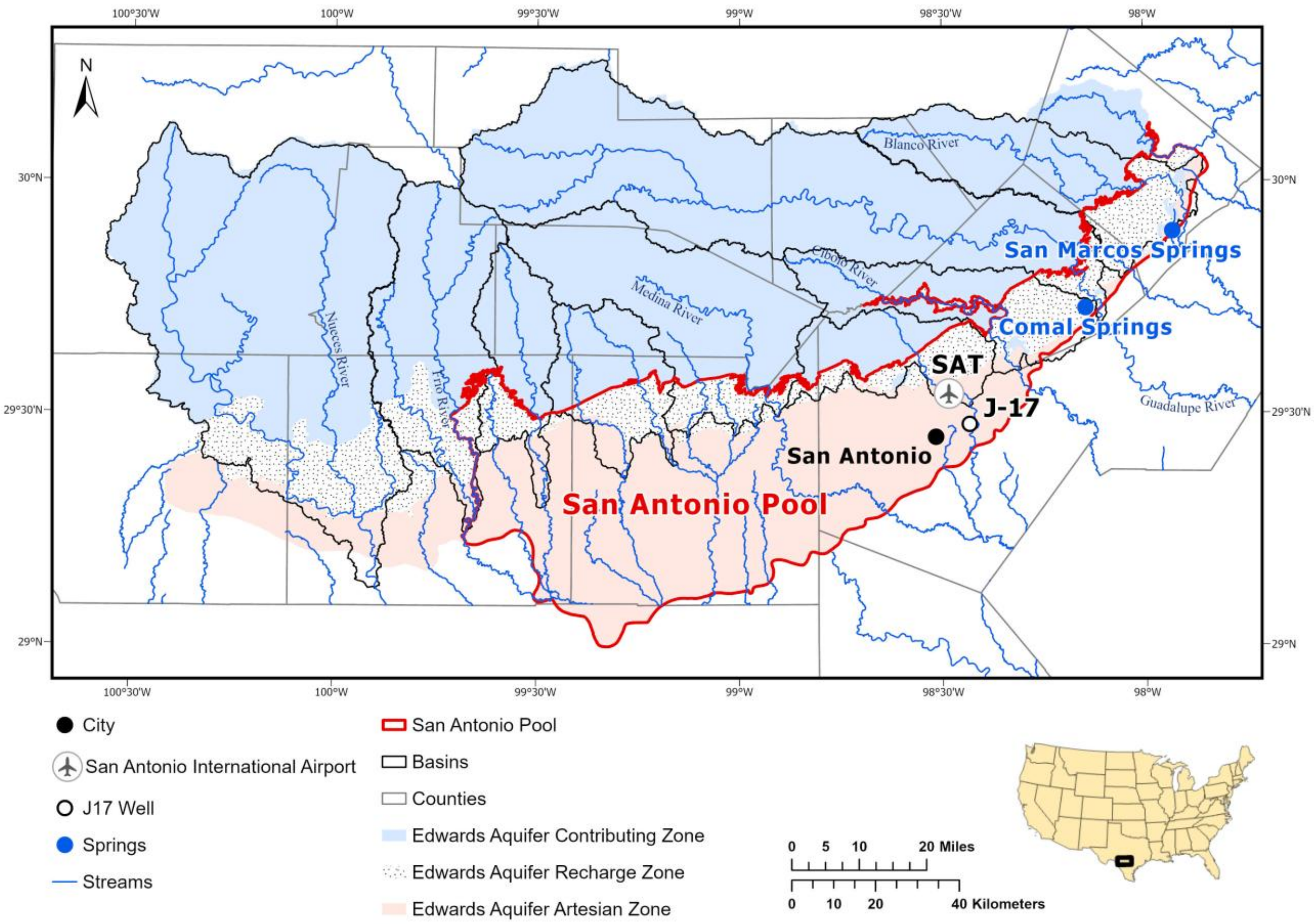

Figure 1: The karstic Edwards Aquifer Region (EAR) in south-central Texas comprises three distinct zones from north to south: the Contributing zone, Recharge zone, and Artesian zone. The J-17 index well serves as a key indicator of hydrological conditions within the San Antonio Pool of the EAR. Comal and San Marcos Springs, the two largest springs in south-central Texas, provide critical habitat for several threatened and endangered endemic species. The San Antonio International Airport (SAT) hosts the region's longest continuous climate record.

The region has experienced intense hydrological droughts, notably in the 1950s, 2010s, and an ongoing event since 2022[11]. Since the early 2000s, the EAA has implemented drought mitigation strategies, including mandatory pumping reductions triggered by groundwater levels and/or spring flows and incentives such as monetarily compensated forbearance agreements for agricultural water-use reductions[12]. In addition, water transfers from an aquifer storage and recovery system are applied during significant droughts to alleviate stress on the aquifer and maintain the minimum required spring flows.

In addition to meteorological factors, researchers have investigated various indicators such as turbidity[13,14], electrical conductivity[15], and tracer data[16] to assess flow paths, water quality, and predict spring flows ($Q_s$) or groundwater levels ($GWL$). While these variables can improve model accuracy, their use is often constrained by limited data availability, intensive field monitoring requirements, and the lack of long-term records. To maintain model simplicity and ensure broad applicability, this study focuses on readily available meteorological inputs, including precipitation and temperature.

Daily precipitation ($P$), and maximum and minimum temperature ($T_{max}$ and $T_{min}$) records from September 1946 to April 2025 were obtained from the San Antonio International Airport (SAT) meteorological station through the NOAA database. Daily $Q_s$ and $GWL$ for the same period were acquired from the USGS and EAA, respectively. Since early 2006, the EAA has implemented a Critical Period Management (CPM) plan as part of drought mitigation strategies to protect the aquifer and the aquatic habitats of several endangered and threatened species during drought conditions. The CPM mandates pumping reductions of 20-44% in the San Antonio Pool of the aquifer, based on $GWL$ at the J-17 index well and flow rates in Comal and San Marcos Springs, relative to five critical stages (CS1-CS5)[17].

For AI model development, the tabular dataset consisting of $P$, $T_{max}$, $T_{min}$, $GWL$, and $Q_s$ was divided into training (September 1946 to December 2009), validation (January 2010 to June 2016), and testing (July 2016 to April 2025) periods. The data were resampled to a weekly time step by summing precipitation and calculating temperature extremes. To analyze temporal variability, the time series of spring flows at Comal Springs and groundwater levels at the J-17 well were decomposed into trend, seasonal, and residual components.

## 3. Methods

### 3.1 Predictor-Predictand Relationships in ETMs and DLMs

The predictor–predictand relationship in the ETMs is defined as follows

$$Q_{t+l} = f\left(\{\mathrm{Pre}_{t-i}, T_{\mathrm{max},t-i}, T_{\mathrm{min},t-i}\}_{i=l}^{l+11},\ \sin\left(\frac{2\pi\cdot \mathrm{month}_t}{12}\right),\ Q_t\right) \quad [1]$$

The predictor–predictand relationship in the DLMs is defined as

$$Q_{t+l} = f\left(\{\mathrm{Pre}_{t-i}, T_{\mathrm{max},t-i}, T_{\mathrm{min},t-i}\}_{i=0}^{t_s-1},\ \sin\left(\frac{2\pi\cdot \mathrm{month}_t}{12}\right),\ Q_t\right) \quad [2]$$

Where:

- $f(\cdot)$ is the model function, which may be ETMs (e.g., XGBoost, ERT) or a DLM (e.g., LSTM, CNN, Transformer),
- $Q_{t+l}$ is the target variable, representing $GWL$ or $Q_s$ at $t + l$ weeks, $l \in \{1,2, \ldots ,12\}$,
- $Q_t$ is the observed value at time step $t$, representing the current system state (e.g., $GWL$ or $Q_s$),
- $\mathrm{Pre}_{t-i}$, $T_{\mathrm{max},t-i}$, and $T_{\mathrm{min},t-i}$ are the meteorological predictors, $P$, $Tmin$, and $T_{max}$ at time lag $i$,
- $t_s$ is the look-back window length (only applicable for deep learning models), i.e., the number of past weeks considered as input for the sequences, and

- $\sin\left(\frac{2\pi\cdot \text{month}_t}{12}\right)$ is the seasonal sine transformation of the calendar month $t$, used to capture intra-annual cyclicality.

The hyperparameter grids for ETMs and DLMs are summarized in Tables S1 and S2. Model performance was evaluated using $R^2$, $KGE$, and $RMSE$ on training, validation, and testing periods. The best configurations were identified based on validation results and subsequently retrained on the combined training–validation set for final testing, and agentic deployment & forecasting.

### 3.2 Model families and feature geometry

The benchmark included extreme gradient boosting (XGBoost), extremely randomized trees (ERT), long short-term memory networks (LSTM), convolutional neural networks (CNN), and Transformers. The comparison was deliberately cross-paradigm: XGBoost and ERT operate through recursive partitioning of tabular predictors, whereas the neural models learn temporal representations from sequential inputs (Table 2).

Table 2: Model and feature designs

| Model group | Families | Feature representation | Operational implication |
|---|---|---|---|
| Tree ensembles | XGBoost, ERT | Twelve climate slots for each of three variables, lagged hydrologic state, and seasonality (38 features) | Can accept prospective weather values in future-facing climate slots |
| Sequential networks | LSTM, CNN, Transformer | One lagged value per climate variable, lagged hydrologic state, and seasonality, reshaped into sequences | Existing trained schema contains no independent future-weather slot |

### 3.3 Deployment-aware weather inputs

At forecast time, tree-model feature rows include both historical and future-relative climate slots. The first future week is retrieved from the National Weather Service (NWS) when a valid forecast is available. If retrieval fails, a four-week trailing persistence estimate is used. Longer future-relative slots currently use the same persistence baseline because the operational point forecast extends only about seven days. In contrast, the trained sequential feature schema resolves its climate inputs to the last observed state and therefore cannot consume independent prospective weather without changing and retraining the model. This distinction exposes a deployment issue that is often omitted from hydrologic model comparisons. During training, future-relative climate slots can be populated from the completed historical record; during live inference, those quantities are uncertain and must be forecast or approximated. The present study therefore distinguishes retrospective predictive skill from operational input realism. The deployed forecast ledger is used to measure the consequences prospectively rather than assuming that high retrospective skill transfers unchanged to production.

### 3.4 Evaluation metrics

Performance was evaluated using the coefficient of determination ($R^2$), root-mean-square error (RMSE), and Kling-Gupta efficiency (KGE). KGE was the primary model-selection metric because it jointly evaluates correlation, variability, and mean bias. For every site and horizon, the operational model was selected by deterministic maximization of KGE; predictions from the five families were not blended.

### 3.5 Critical-stage classification

Continuous predictions were translated into threshold events using five escalating Critical Stages (Table 3). A stage is breached when the weekly minimum falls below its threshold; when several thresholds are crossed, the deepest stage governs interpretation. These classifications were evaluated by agreement between predicted and observed threshold status. The dashboard output is an early-warning indicator and not an official regulatory declaration, which follows separate certified procedures.

Table 3: Critical-stage thresholds used for operational evaluation.

| Stage | Comal discharge ($m^3 s^{-1}$) | J-17 elevation (m) |
|---|---|---|
| CS1 | 6.371 | 201.168 |
| CS2 | 5.663 | 198.120 |
| CS3 | 4.248 | 195.072 |
| CS4 | 2.832 | 192.024 |
| CS5 | 1.274 | 190.500 |

### 3.6 Agentic Artificial Intelligence Framework

This system integrates the validated models (Core Models) for Comal Springs and J-17 well into a live, forecasting system using Agentic AI Framework (summarized in Table 4). No part of the core model's science was re implemented or altered. The automatic framework uses the already trained 120 models to generate weekly forecasts for both stations. Each weekly run fetches the newest available reading for both indicators and for climate. Comal Springs discharge is retrieved from the USGS NWIS daily data. The J-17 well level, collected using EAA data. Climate data: daily maximum and minimum temperature and precipitation is retrieved from NOAA's Climate Data Online API for the San Antonio Airport Station.

**Autonomous AI Agents:** The weekly execution is driven by five cooperating agents coordinated by a centralized orchestrator, shifting away from a single monolithic script. Each agent operates with a fixed objective and a defined toolset, autonomously determining its sequence of execution steps and fallback strategies during failures. By design, agents never alter operational thresholds, model selection metrics, or archival logic, all of which remain under strict, deterministic algorithmic control. Within this framework, the Data Ingestion Agent fetches incoming real-time data streams to update the weekly historical archive, while the Model Orchestration Agent evaluates historical accuracy metrics to dynamically assign the top-performing model to each of the 24 operational slots (spanning 12 forecasting horizons across two sites). Concurrently, the Prediction and Monitoring Agent generates weekly forecasts, audits predictions against five critical stage triggers to handle anomaly alerts, and benchmarks past forecasts against newly received empirical observations. Finally, the Literature Search Agent queries academic and operational repositories to surface recent scientific research relevant to live aquifer conditions, enabling the Report Generation Agent to aggregate all outputs into a comprehensive, timestamped weekly operational report.

**Forecast Weather Data Acquisition :**Initially, every ensemble model forecasting horizon relied on a four-week rolling mean persistence estimate in place of prospective meteorological data. Conversely, sequential models utilize a single shift-based lag feature per variable that consistently resolves to the last known empirical observation regardless of the forecasting horizon, rendering them immune to future weather data dependencies. Automated meteorological ingestion was subsequently introduced exclusively for the one-week-ahead horizon. At the initialization of each execution, a seven-day forecast is retrieved from the NWS grid point API for the San Antonio International Airport location. These data are then aggregated into a standard Monday-to-Sunday weekly window, where temperatures are averaged and total precipitation is summed; this processed matrix is then substituted directly into that specific horizon's ensemble model climate feature vector. If this real-time API request fails, the pipeline automatically falls back to the baseline four-week persistence estimate. Extended forecasting horizons spanning weeks 2 through 12 continue to rely entirely on the persistence baseline, as the operational NWS forecast product only extends approximately seven days forward.

**Software Architecture with Integrated Agentic AI Framework:** Using the validated AI models, the agentic workflow is integrated into the software system via a Streamlit web application capable of running on a local machine or a cloud server. The interface features a persistent status sidebar that monitors site cache availability across seven core tabs: (1) Dashboard presents the current weekly status and 1-12-week ahead forecast for both sites, including groundwater levels at an index well and flow at springs; (2) Run Workflow executes the five-agent weekly pipeline with options for site selection, forced retraining, dry-run mode, and execution history tracking; (3) Run Models allows independent model evaluation against the current data cache outside the agent pipeline; (4) Report indexes and displays all weekly reports; (5) Model Metrics details the validation performance matrices across all model families and forecast horizons; (6) Forecast Verification serves as a dynamic scorecard tracking historical forecasts against realized observations; and (7) Settings manages API keys, data tokens, and the dual parameters governing retraining logic.

Table 4: Agent roles and scope. "LLM-Driven" = the language model decides which tool to call or what text to produce. "Gate on Forecast" = whether LLM output can alter a numeric forecast value. No agent can alter a computed forecast, a model selection index, a threshold, or an archival entry.

| Agent | Primary Function | LLM-Driven? | Gate on Forecast? |
|---|---|---|---|
| Data Ingestion Agent | Fetch USGS / EAA / NOAA data; update weekly parquet cache | Yes (tool selection) | No |

| Agent | Primary Function | LLM-Driven? | Gate on Forecast? |
|---|---|---|---|
| Model Orchestration Agent | Assign best model per slot (argmax over rolling KGE); trigger retraining | Partial (retraining decision) | No (selection = deterministic) |
| Prediction & Monitoring Agent | Run inference on 24 slots; alert on CS threshold breach; verify past forecasts | Yes (alert prose only) | No (inference = deterministic) |
| Literature Search Agent | Query EarthArXiv, arXiv for relevant recent work | Yes (query design) | No |
| Report Generation Agent | Aggregate outputs into timestamped weekly Markdown/HTML report | Yes (text generation) | No |

* Scientific authority remains outside the language model. Unit conversion, temporal aggregation, feature construction, KGE calculation, model selection, inference, threshold detection, and ledger scoring are deterministic Python operations. Language-model decisions are limited to bounded tool selection, retraining recommendations, literature synthesis, and prose generation. This boundary is central to the proposed notion of auditable agentic hydroinformatics: autonomy coordinates work but does not make the numerical provenance of a forecast irreducible.

### 3.7 Prospective verification

Every forecast is recorded at issue time using a unique combination of site, run date, and lead time. After the forecast date has passed and the corresponding weekly observation becomes available, the record is populated with observed value, signed error, absolute percentage error, and verification date. Verified records are protected from overwrite. This append-and-verify design enables convergence analysis, lead-specific accuracy tracking, bias monitoring, and threshold hit rates using forecasts that genuinely existed before the outcome.

## 4. Results

**Study design at a glance** | Five AI models · Two hydrologic targets (Comal Springs $Q_s$; J-17 *GWL*) · 12 forecast lead times (Weeks 1–12) · ~79 years of hydroclimatic records (1946–2025) · Test period: July 2016 – April 2025 · Evaluation: $R^2$, KGE, RMSE, and Critical Stage (CS1–CS4) threshold accuracy.

### 4.1 Experimental Design and Validation Integrity

**Leakage-controlled multi-step forecasting** *Every forecast horizon depends strictly on observations available at or before the forecast origin. No future information enters model inputs during evaluation. This design makes the performance gap between model classes unambiguous.*

Models were evaluated on a held-out test period spanning July 2016 to April 2025 - a period that encompasses a severe ongoing drought (post-2022) and pronounced interannual variability, providing a demanding test of model stability across diverse aquifer states. The training set (September 1946 – December 2009) and validation set (January 2010 – June 2016) were temporally non-overlapping with the test set.

For Ensemble Tree-Based Models (ETMs), lag features spanning lags $l$ to $l$+11 weeks are used at each forecast horizon, drawing exclusively on observations at or before time $t$. For Deep Learning Models (DLMs), a sliding window of optimized look-back length *ts* is applied; model-generated lagged values substitute for unobserved future values at longer horizons. Both ETM and DLM classes therefore face the same informational constraint at extended lead times. The observed superiority of XGBoost cannot be attributed to informational advantages - it reflects genuine differences in how each model class handles accumulated prediction uncertainty.

**Long historical record on a productive karst system** *79 years of records (1946–2025) span multiple severe droughts (1950s, 2010s, 2022–present) and recharge extremes, providing a rigorous out-of-sample evaluation across diverse aquifer states.*

Time-series decomposition of spring flow and groundwater level data confirmed no persistent long-term trend over the 79-year record, indicating that the aquifer has largely oscillated around a stable mean. However, after 2000, low-flow extremes ($< -3$ standard deviations) became dominant and interannual variability increased — an operational context that places particular stress on models' ability to capture extreme states, making the Critical Stage evaluation especially relevant.

### 4.2 Comal Springs Spring Flow Predictions ($Q_s$)

**Unified multi-horizon, multi-model comparison** *Five model families are evaluated at every one of 12 lead times for both spring flow and groundwater level.*

#### 4.2.1 Short-Term (Weeks 1–4)

**Evidence that simpler models outperform deep learning** *XGBoost leads all models at every short-horizon metric and maintains this advantage through all 12 lead times. The gap is not explained by differences in tuning effort - all models received equivalent hyperparameter optimization (*Table 5 *and Figure S1).*

XGBoost achieved the highest accuracy in Week 1 ($R^2 = 0.99$, KGE = 0.99, RMSE = 0.22 m³/s) and sustained $R^2 \geq 0.97$ and KGE ≥ 0.97 through Week 4 with RMSE ≤ 0.47 m³/s. ERT also performed strongly in Weeks 1–2 ($R^2 = 0.99$, KGE = 0.98, RMSE = 0.28 m³/s), followed by a modest decline in Week 4 ($R^2 = 0.94$, KGE = 0.95, RMSE = 0.68 m³/s).

Among DLMs, CNN closely followed observed dynamics ($R^2$ ranging from 0.97 in Week 1 to 0.95 in Week 4; RMSE from 0.53 to 0.65 m³/s). LSTM showed weaker early skill (Week 1: $R^2 = 0.94$, KGE = 0.83, RMSE = 0.71 m³/s) but improved by Week 4 ($R^2 = 0.97$, KGE = 0.93, RMSE = 0.53 m³/s). Transformer remained stable through Week 3 ($R^2 \approx 0.96$–$0.97$; KGE ≈ 0.96) but declined in Week 4 ($R^2 = 0.94$, KGE = 0.89, RMSE = 0.70 m³/s).

Table 5: Comal Springs short-term performance metrics (test period). RMSE in m³/s.

| Model | Week 1 R² | Week 1 KGE | Week 1 RMSE (m³/s) | Week 4 R² | Week 4 KGE | Week 4 RMSE (m³/s) |
|---|---|---|---|---|---|---|
| **XGBoost** | 0.99 | 0.99 | 0.22 | 0.97 | 0.97 | 0.47 |

| Model | Week 1 $R^2$ | Week 1 KGE | Week 1 RMSE ($m^3/s$) | Week 4 $R^2$ | Week 4 KGE | Week 4 RMSE ($m^3/s$) |
|---|---|---|---|---|---|---|
| **ERT** | 0.99 | 0.98 | 0.28 | 0.94 | 0.95 | 0.68 |
| **CNN** | 0.97 | 0.94–0.96 | 0.53 | 0.95 | 0.94–0.96 | 0.65 |
| **LSTM** | 0.94 | 0.83 | 0.71 | 0.97 | 0.93 | 0.53 |
| **Transformer** | 0.96–0.97 | 0.96 | 0.53–0.58 | 0.94 | 0.89 | 0.70 |

**Operational drought-stage evaluation** *Models are assessed against the Edwards Aquifer Critical Stage (CS) thresholds that trigger mandatory pumping reductions of 20–44%. Agreement at CS4 is the hardest test: only 59 events occurred in the test period.*

Critical Stage accuracy in the short-term window highlighted pronounced differences at low-flow thresholds. XGBoost consistently exceeded 96% agreement at CS1–CS3 in Weeks 1–3 and retained 83.9% accuracy at CS4 in Week 4, demonstrating robustness across both normal and critical low-flow conditions. ERT showed comparable skill in earlier weeks but declined to 80.0% at CS4 in Week 4 *(Figure S2)*.

LSTM exhibited the largest discrepancies, with CS4 accuracy as low as 10.2% in Week 1 - despite maintaining >90% accuracy at CS1–CS2 in all weeks. This pattern reflects the tendency of LSTM to overpredict low flows (see Section 5), which is operationally critical because false positives at CS4 would lead to premature relaxation of pumping restrictions. CNN showed competitive performance at CS1–CS3 in Weeks 1–2 (100% at CS3 in Week 2) but dropped to ~55% at CS4 in Week 4. Transformer maintained stability above 94% for CS1–CS3 across all weeks but underperformed at CS4 (33.9% in Week 4) – (see *Figure S2)*.

### 4.2.2 Medium-Term (Weeks 5–8)

**Mechanistic explanation: spectral bias and peak smoothing** *CNN collapse at Week 7 and LSTM instability exemplify the spectral bias of neural networks: their preference for smooth, low-frequency functions conflicts with abrupt, threshold-driven karst recharge responses.*

XGBoost remained the most stable across Weeks 5–8, maintaining $R^2$ = 0.96–0.97 and RMSE = 0.52–0.59 $m^3/s$. ERT retained reasonable skill (Week 5: $R^2$ = 0.93, KGE = 0.93, RMSE = 0.75 $m^3/s$) but declined to $R^2$ = 0.91, KGE = 0.90, RMSE = 0.90 $m^3/s$ by Week 8 and increasingly struggled with low-flow conditions *(*Table 6 *and Figure S3)*.

CNN retained moderate skill at Week 5 ($R^2$ = 0.96, KGE = 0.97, RMSE = 0.59 $m^3/s$) but **collapsed entirely in Week 7, showing near-zero predictive skill** - a failure with direct operational implications, as no alert or forecast would have been produced at that horizon. Partial recovery occurred in Week 8 ($R^2$ = 0.93, RMSE = 0.79 $m^3/s$), but high flows remained consistently under-predicted. LSTM showed declining stability: $R^2$ = 0.93, RMSE = 0.80 $m^3/s$ in Week 5, with sharp performance drops in Week 7 (pronounced under-prediction of high flows and over-prediction of low flows), before partial recovery in Week 8 ($R^2$ = 0.95, RMSE = 0.63 $m^3/s$). Transformer maintained moderate agreement ($R^2$ = 0.95, KGE = 0.92 in Week 5) but lagged during high-flow transitions *(*Table 6 *and Figure S3)*.

Table 6: Comal Springs medium-term performance metrics. RMSE in m³/s. "FAIL" = near-zero $R^2$; model-generated forecasts cannot be relied upon operationally.

| Model | Wk 5 $R^2$ | Wk 5 RMSE | Wk 6 $R^2$ | Wk 6 RMSE | Wk 7 $R^2$ | Wk 7 RMSE | Wk 8 $R^2$ | Wk 8 RMSE |
|---|---|---|---|---|---|---|---|---|
| **XGBoost** | 0.97 | 0.52 | 0.97 | 0.54 | 0.96 | 0.57 | 0.96 | 0.59 |
| **ERT** | 0.93 | 0.75 | 0.92 | 0.80 | 0.91 | 0.86 | 0.91 | 0.90 |
| **CNN** | 0.96 | 0.59 | 0.92 | 0.74 | ~0.00 | FAIL | 0.93 | 0.79 |
| **LSTM** | 0.93 | 0.80 | 0.94 | 0.68 | unstable | ↑↑ | 0.95 | 0.63 |
| **Transformer** | 0.95 | 0.62 | 0.94 | 0.66 | 0.93 | 0.71 | 0.93 | 0.72 |

Medium-term CS accuracy: XGBoost ranged from 94.3–98.1% at CS1–CS3 in Weeks 5–6, declining to 68.2% at CS4 by Week 8. CNN failed at all CS levels in Week 7 before partial recovery. Transformer showed the greatest stability among DLMs (92.6–96.2% at CS1–CS3), but CS4 accuracy fluctuated from 25% (Week 6) to 80% (Week 7). LSTM reached as low as 9.2% at CS4 in Week 7 *(Figure S4)*.

### 4.2.3 Long-Term (Weeks 9–12)

**Horizon-dependent DLM degradation** *All DLMs show increasing error accumulation and instability at Weeks 9–12. XGBoost's ability to maintain $R^2 \geq 0.94$ across this entire window - while deep learning models collapse or degrade - demonstrates the consequence of inductive bias mismatch at extended horizons (*Table 7 *and Figures S5 and S6).*

XGBoost remained the most reliable at extended horizons, maintaining $R^2$ = 0.95–0.96, KGE = 0.92–0.93, and RMSE = 0.60–0.64 m³/s. ERT showed a more pronounced decline, with $R^2$ decreasing to 0.87 and RMSE exceeding 1.0 m³/s in Week 12, with systematic overprediction during low-flow events.

Among DLMs, CNN performed best in this horizon ($R^2$ mostly above 0.90), though it smoothed peaks and underestimated variability, particularly during low-flow conditions. Transformer produced smoother forecasts ($R^2$ = 0.90–0.94; RMSE = 0.70–0.92 m³/s) but consistently overpredicted low flows. LSTM was the least stable, with $R^2$ = 0.91–0.93 and pronounced phase mismatches at both extremes, reflecting limited capacity to capture long-term dependencies.

Table 7: Comal Springs long-term performance summary. RMSE in m³/s. CS4 agreement = % of observed CS4 events correctly identified.

| Model | $R^2$ Range (Wks 9–12) | KGE Range | RMSE Range (m³/s) | CS1–CS3 Agreement | CS4 Agreement |
|---|---|---|---|---|---|
| **XGBoost** | 0.94–0.96 | 0.92–0.93 | 0.60–0.64 | >90% | 52.9–59.7% |
| **ERT** | 0.87–0.92 | 0.85–0.91 | 0.80–1.01 | <82% | 35.8–47.7% |
| **CNN** | 0.90–0.94 | 0.91–0.95 | 0.60–0.92 | >92% | 56.5–76.5% |
| **LSTM** | 0.91–0.93 | 0.88–0.93 | 0.64–0.84 | 90.9–98.8% | 17.4–47% |

| Model | $R^2$ Range (Wks 9–12) | KGE Range | RMSE Range ($m^3/s$) | CS1–CS3 Agreement | CS4 Agreement |
|---|---|---|---|---|---|
| **Transformer** | 0.90–0.94 | 0.88–0.95 | 0.70–0.92 | 90.5–98.8% | 26.1–55.2% |

### 4.3 J-17 Index Well Groundwater Level Predictions (*GWL*)

Results for *GWL* closely mirror those for Comal Springs, reinforcing the generalizability of the model ranking across both hydrologic targets.

#### 4.3.1 Short-Term (Weeks 1–4)

XGBoost achieved $R^2$ = 0.99, KGE = 0.99, and RMSE = 0.49 m in Week 1, maintaining strong performance through Week 4 ($R^2$ = 0.97, KGE = 0.97, RMSE = 0.87 m). ERT followed closely in Week 1 ($R^2$ = 0.99, KGE = 0.98, RMSE = 0.62 m) but declined more sharply to $R^2$ = 0.94, KGE = 0.93, RMSE = 1.37 m in Week 4 *(*Table 8 *and Figure S7)*.

Among DLMs, CNN remained competitive (Week 1: $R^2$ = 0.96, KGE = 0.96, RMSE = 1.03 m; near identical in Week 4), tending to smooth peak values. LSTM started (Week 1: $R^2$ = 0.96, RMSE = 1.08 m) but degraded to $R^2$ = 0.93, RMSE = 1.51 m in Week 4, reflecting damped variability at longer leads. Transformer remained stable in $R^2$ (0.96–0.95 through Weeks 1–4) but exhibited higher RMSE (1.08–1.20 m) and lower KGE (0.98–0.90), indicating amplitude mismatches despite acceptable phase alignment. All DLMs tended to overpredict extreme low-level events *(Figure S7)*.

Table 8: J-17 short-term performance metrics. RMSE in meters.

| Model | Week 1 $R^2$ | Week 1 KGE | Week 1 RMSE (m) | Week 4 $R^2$ | Week 4 KGE | Week 4 RMSE (m) |
|---|---|---|---|---|---|---|
| **XGBoost** | 0.99 | 0.99 | 0.49 | 0.97 | 0.97 | 0.87 |
| **ERT** | 0.99 | 0.98 | 0.62 | 0.94 | 0.93 | 1.37 |
| **CNN** | 0.96 | 0.96 | 1.03 | 0.96 | 0.97 | 1.03 |
| **LSTM** | 0.96 | 0.96 | 1.08 | 0.93 | 0.88 | 1.51 |
| **Transformer** | 0.96 | 0.98 | 1.08 | 0.95 | 0.90 | 1.20 |

CS-level accuracy at J-17 for short lead times reveals the most striking DLM failure in the study: LSTM collapsed to **0.0% at CS4 in Week 4** - correctly identifying zero of the observed critical low-level events at that horizon. This occurred despite maintaining >90% accuracy at CS1–CS2, confirming that aggregate regression metrics ($R^2$) can mask operationally catastrophic failures at drought extremes. XGBoost sustained 97.1–95.2% agreement at CS1–CS3 (Week 1) and 93.0–96.3% in Week 4, with moderate skill at CS4 (90.9% Week 1; 79.5% Week 4) *(Figure S8)*.

#### 4.3.2 Medium-Term (Weeks 5–8)

XGBoost maintained $R^2$ and KGE at 96–97% and 0.95–0.96, respectively, from Week 5 through Week 8, with RMSE increasing only modestly from ~0.97 m to 1.17 m - consistent with gradual, proportional error accumulation rather than structural breakdown. The model accurately reproduced both the amplitude and timing of peak and recession phases, with only slight underprediction near peak *GWL* values.

ERT performance was slightly lower ($R^2$ 0.93–0.90; KGE 0.91–0.88; RMSE 1.51–1.80 m), with overprediction during the falling limb of *GWL* responses and increasing error spread during pronounced fluctuations. Among DLMs, LSTM was the most reliable across the medium-term range, maintaining $R^2$ = 0.97 (Week 5) to 0.95 (Week 8) and KGE = 0.85–0.98, effectively preserving seasonal patterns and directional shifts. CNN declined from $R^2$ = 0.95 to 0.92 and RMSE from 1.26 m to 1.61 m, with both underprediction of peaks and damped oscillations. Transformer showed the greatest variability ($R^2$ 0.90–0.95; KGE 0.84–0.98), with predictions often lagging behind observed peaks and troughs *(Figure S9)*.

CS4 accuracy at J-17 in the medium term exposed the most severe DLM failure of the entire study: **LSTM accuracy at CS4 collapsed to 0% in Week 6** before recovering to 75% in Week 7. Transformer performed near zero at CS3–CS4 in intermediate weeks. ERT deteriorated to just 14% at CS4 through Week 8. In contrast, XGBoost maintained 45–70% agreement at CS4 across all medium-term horizons *(Figure S10)*.

### 4.3.3 Long-Term (Weeks 9–12)

**Catastrophic DLM failure: CNN at Week 9** *CNN produced a negative $R^2$, undefined KGE, and RMSE > 5.8 m at Week 9 (J-17) - essentially outputting a near-constant flat prediction. This illustrates the operational risk of deploying DLMs without horizon-specific reliability testing (*Table 9 *and Figures S11 and S12).*

XGBoost sustained the highest accuracy at extended horizons ($R^2$ = 0.94–0.96, KGE = 0.95–0.96, RMSE = 1.15–1.37 m), preserving trend and amplitude of *GWL* variations with only slight errors during recession phases. ERT showed more pronounced decline ($R^2$ decreasing from 0.89 to 0.85; KGE 0.87 to 0.82; RMSE 1.87–2.14 m), with systematic overprediction of low *GWL* and underestimation of high *GWL*.

CNN suffered a catastrophic failure in Week 9: negative $R^2$, undefined KGE, and RMSE exceeding 5.8 m due to flat (near-constant) predictions. It recovered in subsequent weeks ($R^2$ = 0.92–0.93; KGE = 0.93–0.95) but remained overly smooth and underestimated variability. LSTM was the most stable DLM in this range ($R^2$ = 0.93–0.95; KGE = 0.89–0.91; RMSE = 1.27–1.47 m), consistently capturing timing and overall pattern while damping peak extremes. Transformer maintained $R^2$ = 0.91–0.93 and KGE = 0.90–0.97, but RMSE increased to 1.69 m by Week 12, with growing amplitude bias and lagged response to observed fluctuations.

Table 9: J-17 long-term performance summary. RMSE in meters. † = Week 9 failure values. CS4 = percentage of observed critical low-level events correctly identified.

| Model | $R^2$ Range (Wks 9–12) | KGE Range | RMSE Range (m) | CS1–CS3 Agreement | CS4 Agreement |
|---|---|---|---|---|---|
| **XGBoost** | 0.94–0.96 | 0.95–0.96 | 1.15–1.37 | 94–98% | 42–54% |
| **ERT** | 0.85–0.89 | 0.82–0.87 | 1.87–2.14 | 87–91% | 0% (Wks 10–12) |
| **CNN** | −† / 0.92–0.93 | —† / 0.93–0.95 | 5.8†/ 1.3–1.5 | 0%† / 89–94% | 0%† / 33–47% |

| Model | $R^2$ Range (Wks 9–12) | KGE Range | RMSE Range (m) | CS1–CS3 Agreement | CS4 Agreement |
|---|---|---|---|---|---|
| **LSTM** | 0.93–0.95 | 0.89–0.91 | 1.27–1.47 | 92–98% | 17–47% |
| **Transformer** | 0.91–0.93 | 0.90–0.97 | 1.32–1.69 | 85–95% | 13–40% |

### 4.4 Cross-Site, Cross-Model Synthesis

**Consistent superiority of parsimonious tree models** *XGBoost ranks first on every metric at every horizon for both targets (*Table 10*). This consistency across two independent hydrologic systems (spring discharge and aquifer head) strengthens the inference beyond a single-site observation.*

Table 10: Qualitative cross-model ranking summary across both sites and all horizons. ✓ = best performing. † = excluding Week 9 failure at J-17.

| Metric | XGBoost | ERT | CNN | LSTM | Transformer |
|---|---|---|---|---|---|
| **$R^2$ - Short-term** | 0.97–0.99 ✓ | 0.94–0.99 | 0.95–0.97 | 0.93–0.97 | 0.94–0.97 |
| **$R^2$ - Medium-term** | 0.96–0.97 ✓ | 0.90–0.93 | 0.92–0.96 | 0.93–0.97 | 0.90–0.95 |
| **$R^2$ - Long-term** | 0.94–0.96 ✓ | 0.85–0.92 | 0.90–0.94† | 0.91–0.95 | 0.90–0.93 |
| **CS4 Agreement** | Best overall ✓ | Moderate | Variable (failures) | Poor at CS4 | Variable |
| **Catastrophic failures** | None ✓ | None | Week 7 ($Q_s$) + Week 9 (*GWL*) | CS4 collapse (0%) | CS4 near-zero |
| **RMSE trend (short→long)** | Gradual ✓ | Moderate | High + failures | Moderate + spikes | Moderate + bias |

Three structural patterns emerge consistently across both sites (summarized from *Figures S1* to *S12*):

1. **XGBoost leads all models across all horizons**, demonstrating both the highest accuracy and the most stable degradation trajectory with increasing lead time. The gap relative to the next-best model (ERT at short to medium horizons; LSTM at long horizons) widens with lead time for both $Q_s$ and *GWL*.
2. **DLMs exhibit non-monotonic degradation**, including complete model collapse at specific horizons (CNN at Week 7 for $Q_s$; CNN at Week 9 for *GWL*). $R^2$ and KGE overstate DLM reliability: two models with identical $R^2 = 0.93$ may differ by a factor of 5 in RMSE and by 80 percentage points in CS4 accuracy.
3. **CS4 accuracy is the most discriminating metric**. At CS4 - the threshold triggering the most severe pumping reductions (44%) - XGBoost is the only model to maintain meaningful accuracy (>40%) at all horizons for both sites. ERT collapses to 0% at CS4 for J-17 in the long term; LSTM collapses to 0% at CS4 for J-17 in the medium term; CNN collapses to 0% at all CS levels at Week 7 ($Q_s$) and Week 9 (*GWL*).

### 4.5 Agentic AI Framework: Operational Results

**First agentic-AI framework for operational aquifer forecasting** *This is the first reported system in which cooperating autonomous agents handle the full operational workflow - data retrieval, model selection, prediction generation, CS threshold alerting, forecast verification, literature surveillance, and report generation - for real-time karst aquifer forecasting. The operational interface of the integrated agentic software system is shown in* Figure 2.

#### 4.5.1 Operational Timeline

The prediction modeling framework began generating forecasts on June 8, 2026. Every calendar week from June 14 through November 15, 2026, has at least one forecast on record, with no gaps in target-week coverage. The initial phase (June 8 – August 2, 2026) served as a test and run-in period during which system corrections were made and execution times were irregular. Regular verified operation commenced August 3, 2026. Starting August 17, 2026, Week-1 forecasts incorporated live NWS forecast data in place of the four-week persistence baseline - representing the first operational bridge between weather forecasting and aquifer-state forecasting for this system.

**Dynamic model assignment per site and horizon** *Rather than applying a single fixed production model to all 24 operational slots (2 sites × 12 lead times), the Model Orchestration Agent evaluates KGE independently for each slot and dynamically assigns the best-performing model. This recognizes that model skill can vary by target variable and lead time.*

**Closed-loop adaptive hydroinformatics** *The Prediction & Monitoring Agent archives every forecast and subsequently benchmarks it against realized observations, enabling continuous prospective evaluation. The Model Orchestration Agent can trigger retraining when the rolling KGE falls below threshold, addressing hydroclimatic nonstationarity without manual intervention.*

#### 4.5.2 Prospective Forecast Verification

**Closed-loop forecast verification** *Unlike static retrospective validation - which uses historical data already available to the modeler - these results represent genuine prospective performance: forecasts were locked at issue time, targets were observed later, and errors computed independently. This is the most operationally meaningful performance test possible.*

Across the four weekly operational runs from August 3 through August 24, 2026, the system generated 96 forecasts across the two study sites (12 horizons × 2 sites × 4 weeks). Of these, 12 forecasts have been verified against realized observations (6 per site); the remaining 84 are pending, as their target dates had not yet passed at time of reporting.

A mean absolute percentage error (MAPE) of 4.3% for Comal springs flow and just 0.19% for J-17 groundwater level - with a near-zero mean bias at both sites - confirms that the agentic pipeline maintains, in real-time prospective mode, the predictive accuracy demonstrated in the retrospective validation (Table 11). The negligible mean bias (−0.03 m³/s for $Q_s$; −0.36 m for *GWL*) indicates that no systematic directional error was introduced by the automated data-ingestion or feature-construction pipeline. Expanded results in Tables S3 and S4.

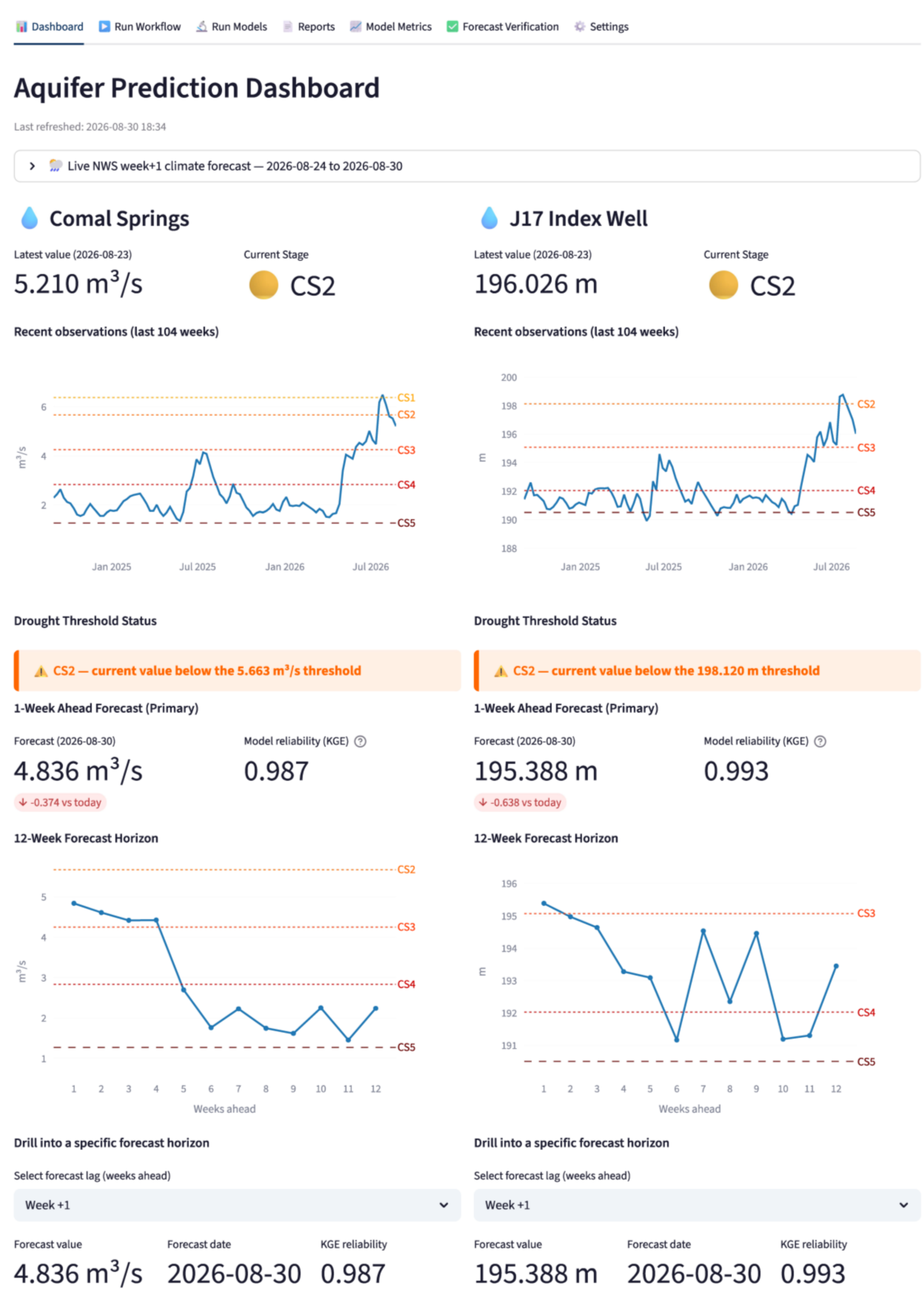

Figure 2: Operational interface of the integrated agentic software system, tracking site telemetry and 1-12-week model forecasts for flow at Comal Springs and groundwater levels at the J-17 index well.

Table 11: Prospective operational forecast verification results (August 3 – August 24, 2026). Mean error and MAPE computed over the first 6 verified forecast weeks per site. Negative mean error = slight underestimation.

| Site | Target Variable | Verified Forecasts | Mean Error | MAPE |
|---|---|---|---|---|
| **Comal Springs** | Spring flow ($Q_s$) | 6 | −0.03 m³/s | 4.3% |
| **J-17 Index Well** | Groundwater level ($GWL$) | 6 | −0.36 m | 0.19% |

## 5. Discussion

### 5.1 Why tree ensembles outperform deep networks here

Three mechanisms explain the observed performance contrast. First, karst responses contain discontinuities and event-driven transitions. Tree ensembles approximate such relationships through piecewise partitions, whereas neural optimization often favors smoother functions[18]. The tendency to underpredict peaks and elevate lows is therefore consistent with model inductive bias rather than merely inadequate tuning.

Second, the predictor space is compact and axis aligned: precipitation, temperature, seasonality, and lagged aquifer state each have a clear physical meaning. Tree methods directly select and threshold these variables. Neural networks must recover the same relationships through distributed weights, leaving much of their representational capacity unused and increasing the opportunity for overfitting or unstable convergence. This interpretation is consistent with broad tabular benchmarks[19] and recent time-series evidence that optimal model class depends on intrinsic data characteristics rather than universal architectural superiority[20].

Third, uncertainty grows with forecast horizon. When future forcing is unavailable, each model must rely more strongly on system persistence and approximated meteorological inputs. Neural smoothing and accumulated uncertainty become most visible at weeks 9-12, especially during scarce low-flow events. Similar horizon-dependent degradation has been reported in groundwater and surface-water forecasting[21,22].

### 5.2 General hydrologic implications

The Edwards Aquifer is not presented merely as a local application. It represents a wider class of operational hydrologic problems with long records but few routinely available predictors. In this regime, the benefit of a model arises from the interaction between its inductive bias and the geometry of the available data. Deep learning may recover an advantage when inputs include gridded remote sensing, spatially distributed soil moisture, dense sensor networks, or other high-dimensional fields. When inputs remain compact and tabular, parsimony can improve accuracy, stability, interpretability, and computational efficiency simultaneously.

The study also shows why hydrologic forecast assessment should extend beyond global error metrics. A model can attain high $R^2$ while smoothing precisely the extremes that drive water-management decisions. KGE penalizes variability and bias, while threshold agreement tests whether errors change the operational conclusion. Together, these metrics offer a more decision-relevant evaluation than correlation alone.

### 5.3 Agentic automation as a scientific workflow

The agent architecture contributes a reproducible operational pattern rather than a claim that language models improve the numerical forecast. Flexible agents are useful where tasks require tool choice, synthesis, or communication. Deterministic code remains preferable where repeatability and traceability dominate. Explicitly separating these roles prevents an LLM from selecting numerical outputs, modifying thresholds, or rewriting historical verification records.

Prospective verification is the most scientifically important operational component. Retrospective tests are necessary but cannot reproduce changing data feeds, stale caches, weather-forecast uncertainty, software failures, or shifting environmental distributions. An immutable issued-forecast record allows these effects to be measured. This converts deployment from an endpoint into a continuing experiment in which model performance, bias, and threshold skill are updated as outcomes mature.

### 5.4 Limitations and next steps

The study uses two indicators from one karst system. The proposed explanation of model-data alignment should be evaluated across aquifers with different memory, recharge, pumping, and climatic regimes. Rare CS4-CS5 events provide limited class support, so high overall threshold agreement does not guarantee sensitivity to the most severe drought states. The initial prospective sample is too small to establish live operational skill.

The largest deployment limitation concerns future meteorological forcing. Only the first future week currently receives an operational NWS point forecast, and only the tree models have feature slots capable of using it. Longer horizons use persistence estimates. A tiered design using NWS forecasts for week 1, Climate Prediction Center outlooks for weeks 2-4, and climatological normals for weeks 5-12 has been prototyped but not incorporated into the results reported here. Its value must be assessed prospectively after deployment.

Finally, agentic execution does not by itself guarantee reliability. Production use requires deterministic completeness checks, durable logs, explicit retry policies, strong data-quality gates, and transparent failure states. The present architecture establishes the scientific trust boundary and verification record; further engineering hardening is required before the system should be treated as fully autonomous regulatory infrastructure.

## 6. Conclusions

This study demonstrates that model complexity is not a reliable proxy for hydrologic forecast skill. Across two indicators and twelve lead times, XGBoost provided the most accurate and operationally stable predictions, while neural architectures increasingly smoothed extremes and occasionally failed at longer horizons. The result is explained by alignment between recursive tree partitioning and the low-dimensional, threshold-driven structure of karst hydroclimatic data.

The broader methodological contribution is a deployment-aware forecasting framework built on four principles: preserve forecast-origin integrity, select models according to data geometry rather than novelty, evaluate decision thresholds in addition to average error, and archive predictions before outcomes occur. Agentic coordination can then automate data acquisition, monitoring, literature retrieval, and reporting without relinquishing deterministic control of scientific calculations.

For operational groundwater systems, the practical lesson is direct: choose the simplest model that reliably represents the available data, expose the assumptions required for live inference, and treat prospective verification as part of the science rather than an optional post-deployment activity.

### CRediT authorship contribution statement

*PL: Investigation, Methodology – Agentic System, Validation, Original draft, Writing – review & editing, Software; CS: Investigation, Methodology – Machine Learning, Validation, Original draft, Writing – review & editing, Software; HB: Conceptualization, Methodology, Data curation, Writing – review & editing, PB: Project administration, Writing – review & editing; DC: Conceptualization, Investigation, Methodology, Writing – review & editing, Software.*

### Declaration of Competing Interest

The authors declare that they have no known competing financial interests or personal relationships that could have appeared to influence the work reported in this paper.

### Data and code availability

All datasets, code, and instructions are available on GitHub:
https://github.com/pramodlekhak/Predictors-and-Orchestrators-Machine-Learning-within-an-Agentic-AI-Harness-for-Forecasting

### Declaration of generative AI and AI-assisted technologies in the manuscript preparation process

During preparation of this manuscript, the authors used OpenAI Codex and Anthropic Claude to support structural revision, synthesis of author-provided materials, language editing, and Word-document preparation. The authors reviewed and edited the content and take full responsibility for the scientific claims and the final submitted version.

### Acknowledgements

*This material is based on work supported by the EAA (award number SAT0004071). Any opinions, findings, conclusions, or recommendations expressed in this publication are those of the authors and do not necessarily reflect the views or regulatory position of the EAA. The authors gratefully acknowledge Dr. Maryam Samimi of the Edwards Aquifer Authority for her assistance in preparing the study area map.*

# *Supplementary - Predictors and Orchestrators: Parsimonious Machine Learning within an Agentic AI Harness for Multi-Horizon Karst Aquifer Forecasting*

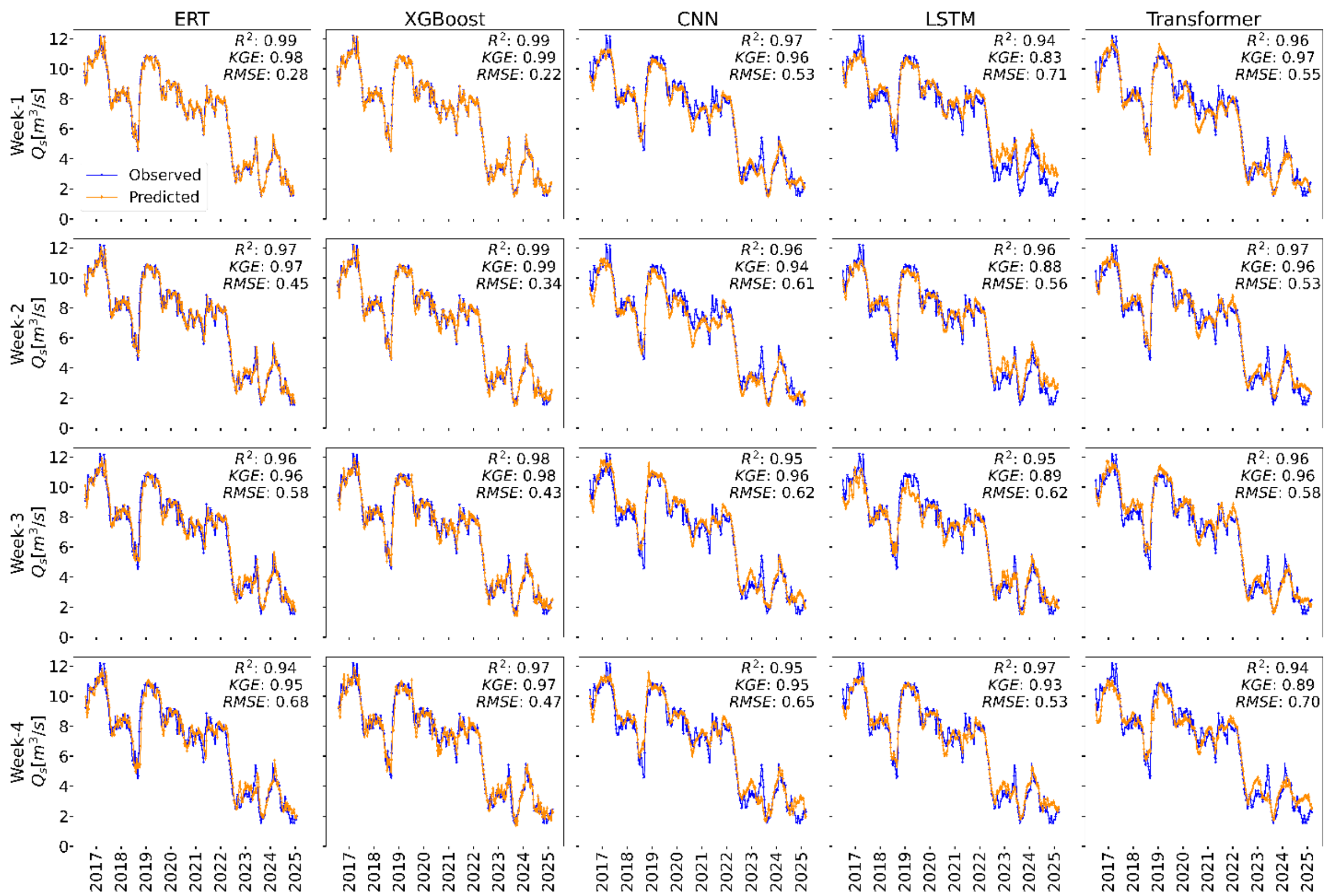

*Figure S3: Short-term (1–4 week) predictions of flows at Comal Springs. Observed (blue) and predicted (orange) spring flows are shown for all models. RMSE is expressed in $m^3/s$.*

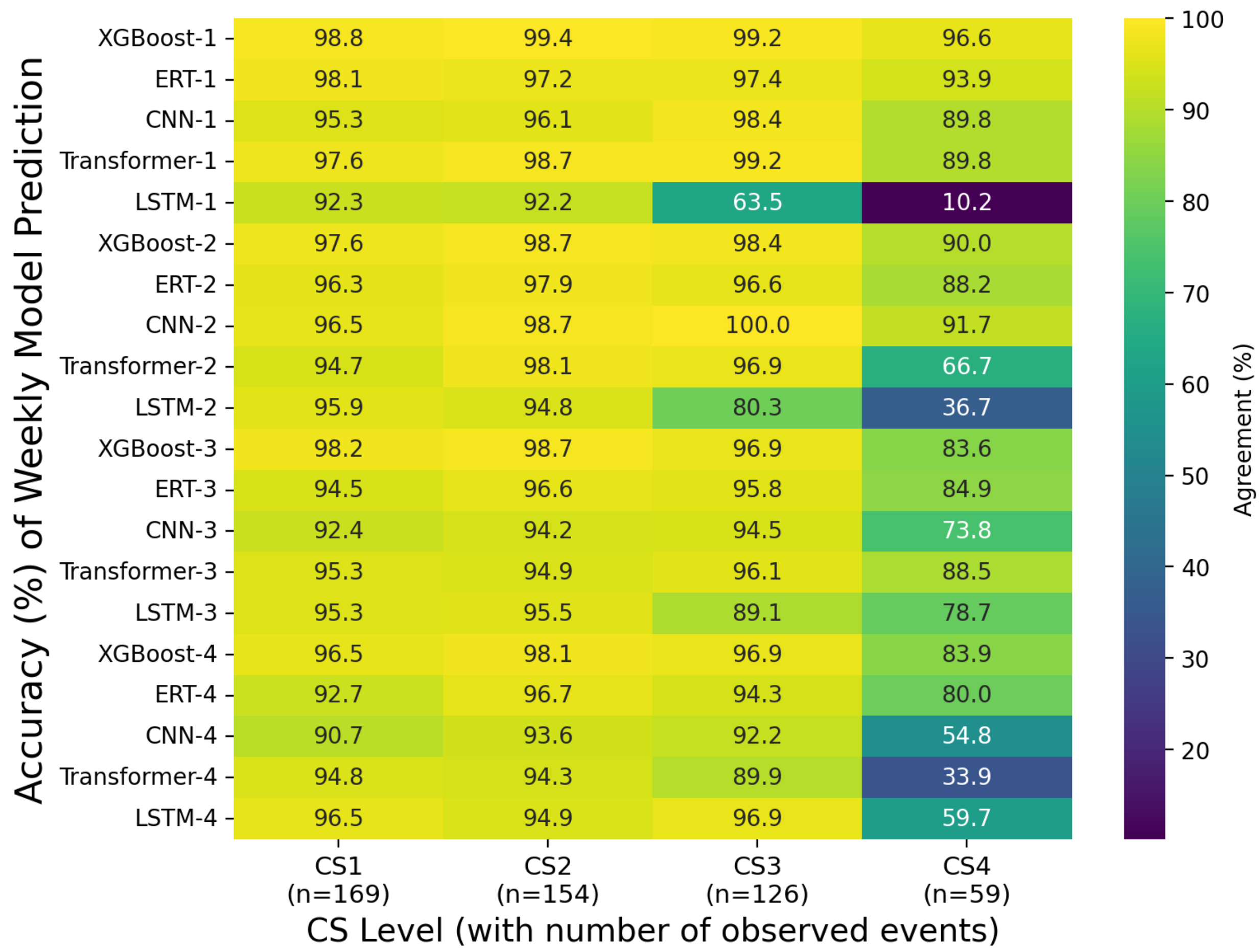

*Figure S4: Short-term (1-4 weeks) flow prediction performance of the AI models with respect to Critical Stage (CS) levels for Comal springs. The numbers on the y-axis represent the week number. Agreement indicates the percentage of times that spring flow values below a given CS threshold in the test dataset were correctly predicted.*

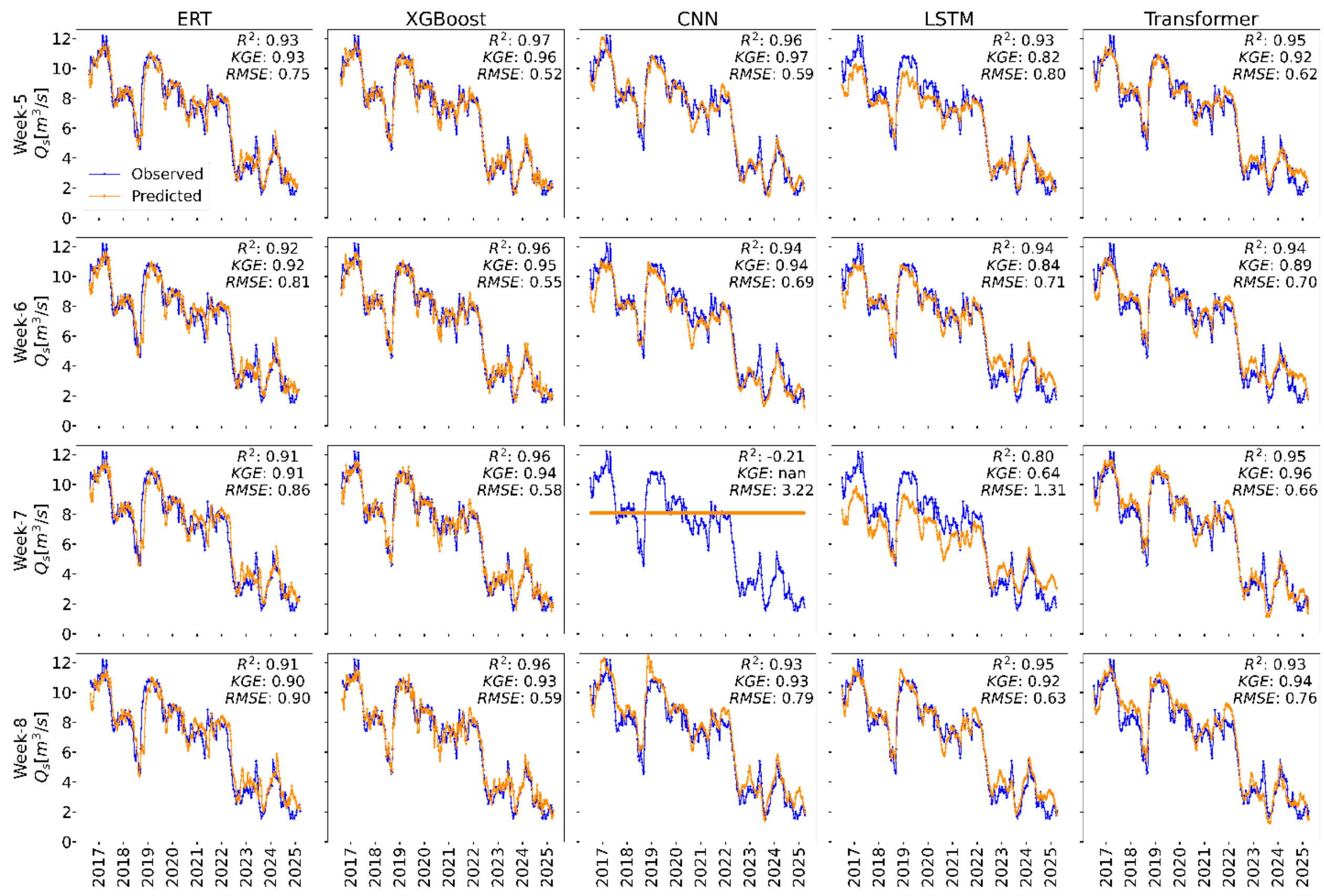

*Figure S5: Medium-term (5-8 week) predictions of flows at Comal Springs. Observed (blue) and predicted (orange) spring flows are shown for all models. RMSE is expressed in m$^3$/s.*

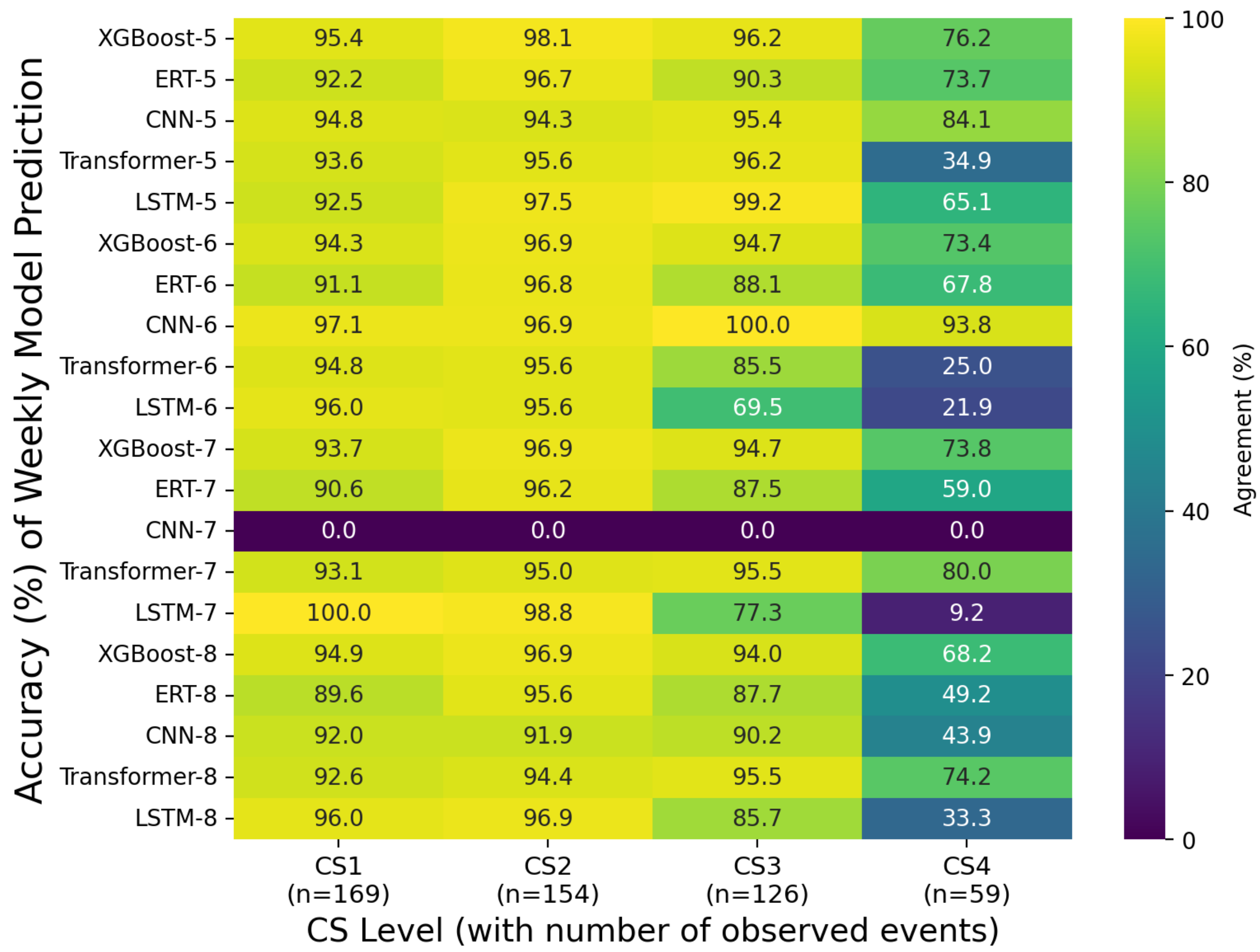

*Figure S6: Medium-term (5-8 weeks) flow prediction performance of the AI models with respect to Critical Stage (CS) levels for Comal Springs. The numbers on the y-axis represent the week number. Agreement indicates the percentage of times that spring flow values below a given CS threshold in the test dataset were correctly predicted.*

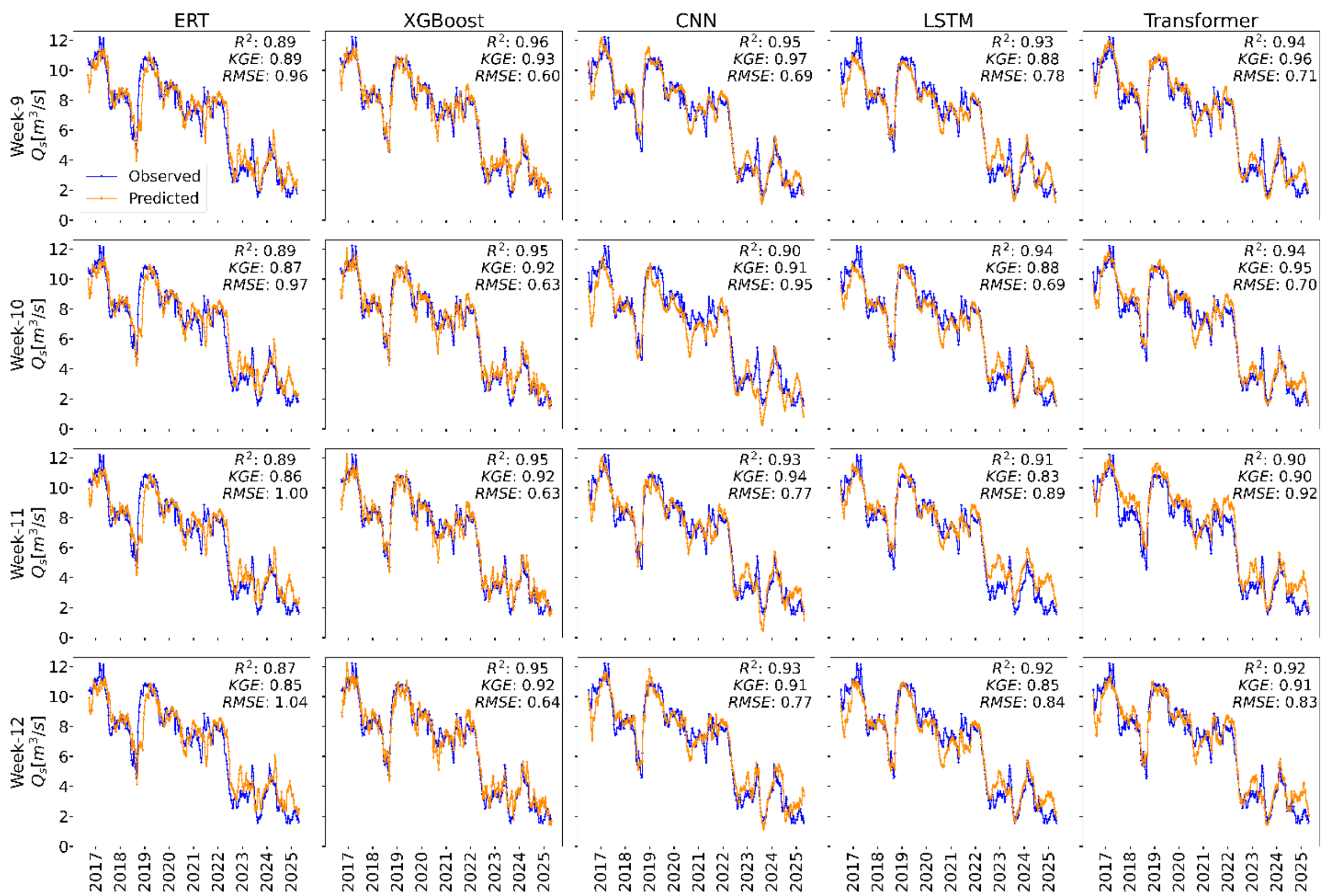

*Figure S7: Long-term (9-12 week) predictions of flows at Comal Springs. Observed (blue) and predicted (orange) spring flows are shown for all models. RMSE is expressed in $m^3/s$.*

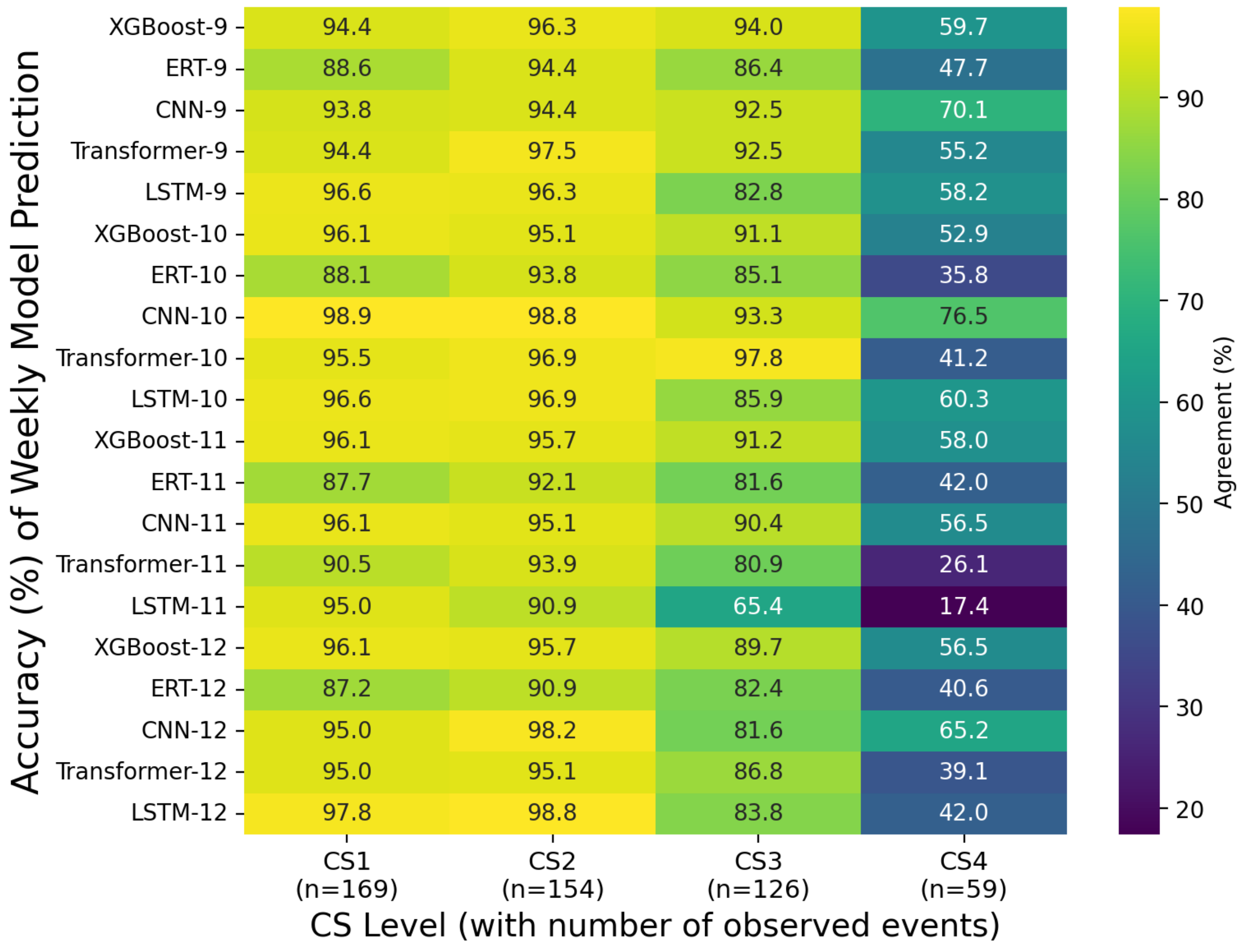

*Figure S8: Long-term (9-12 weeks) flow prediction performance of the AI models with respect to Critical Stage (CS) levels for Comal springs. The numbers on the y-axis represent the week number. Agreement indicates the percentage of times that spring flow values below a given CS threshold in the test dataset were correctly predicted.*

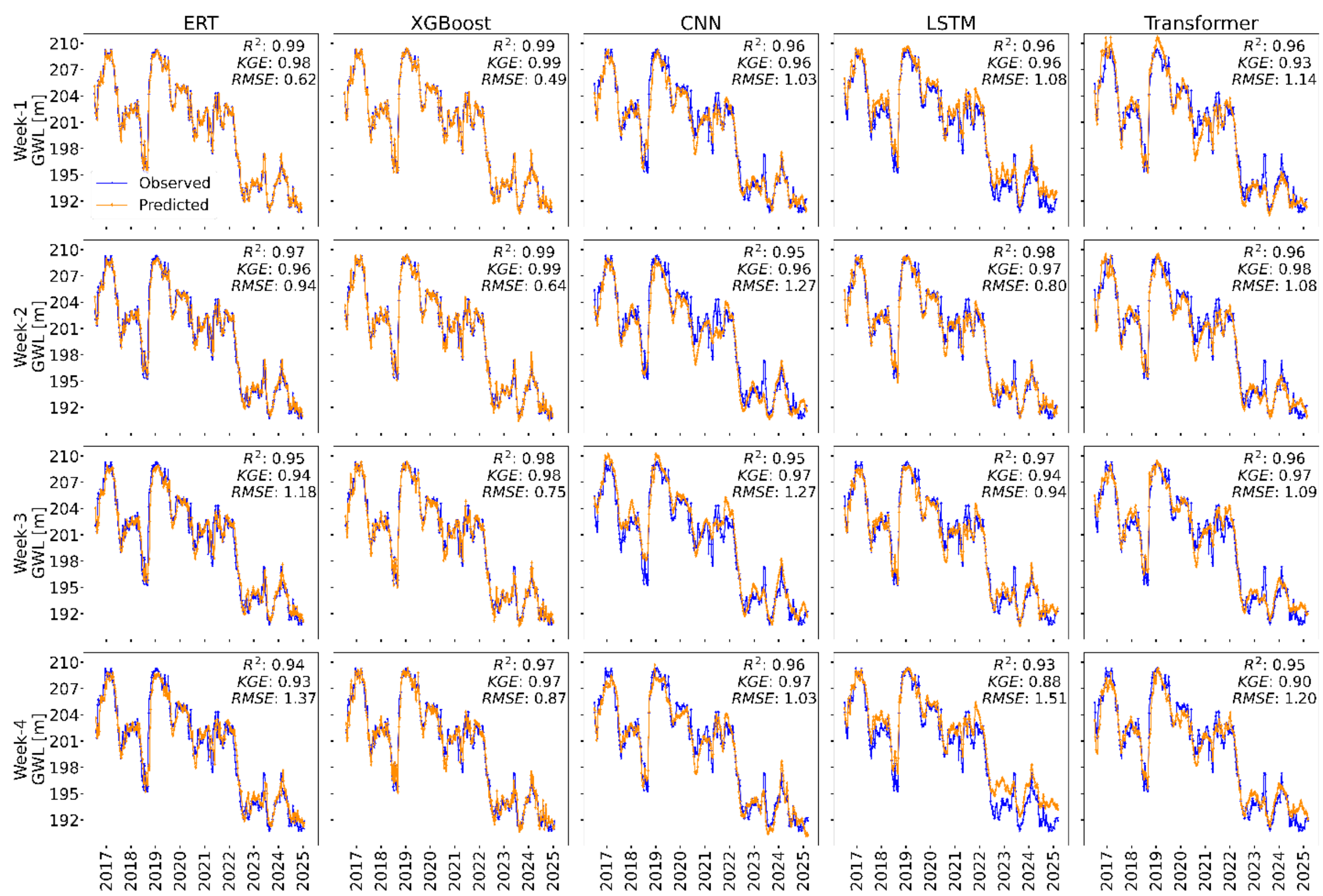

*Figure S9: Short-term (Weeks~1--4) predictions of groundwater levels at the J17 well. Observed (blue) and predicted (orange) groundwater levels are shown for all models. RMSE is expressed in m.*

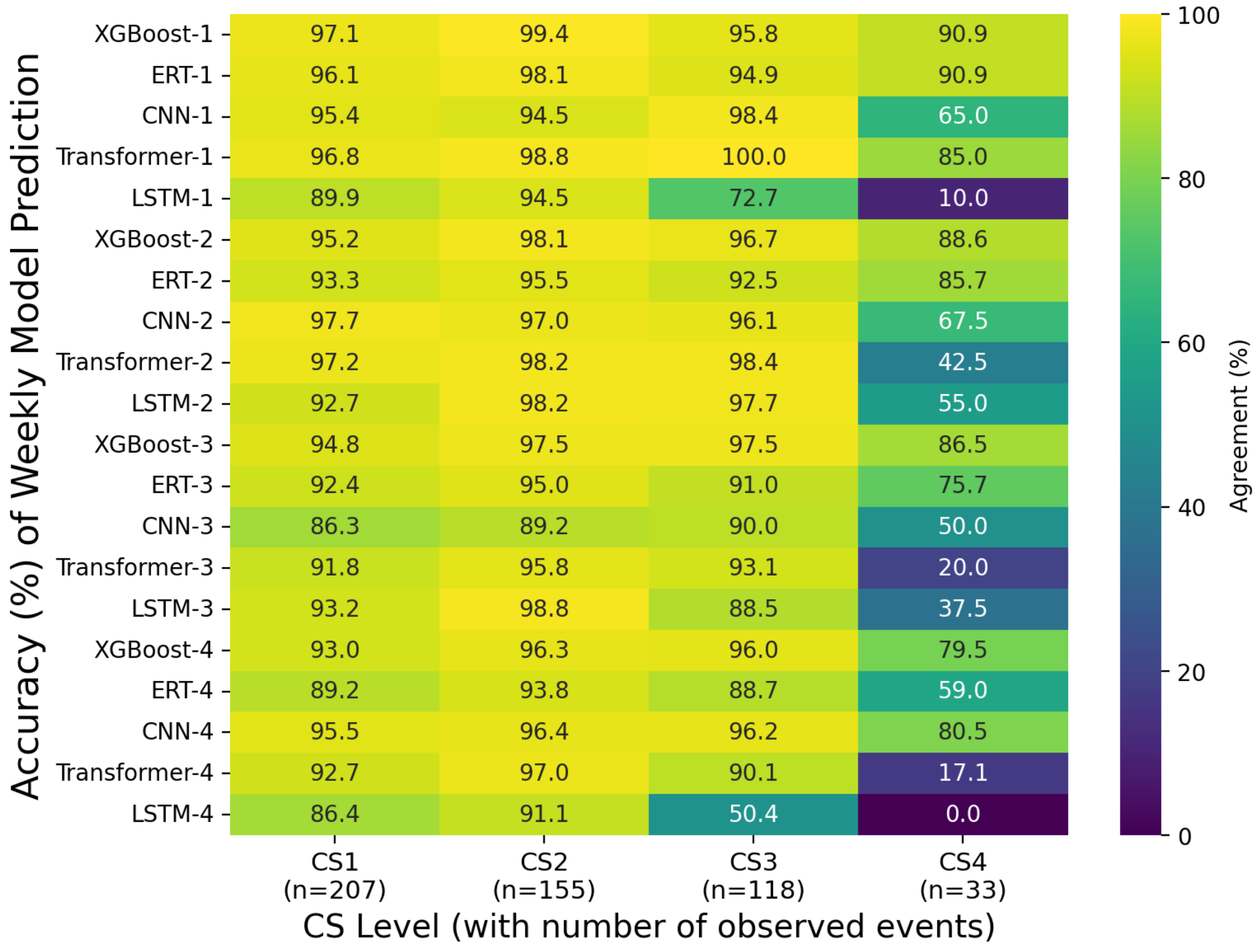

*Figure S10: Short-term (1-4 weeks) groundwater level prediction performance of the AI models with respect to Critical Stage (CS) levels for the J17 well. Agreement indicates the percentage of times that spring flow values below a given CS threshold in the test dataset were correctly predicted.*

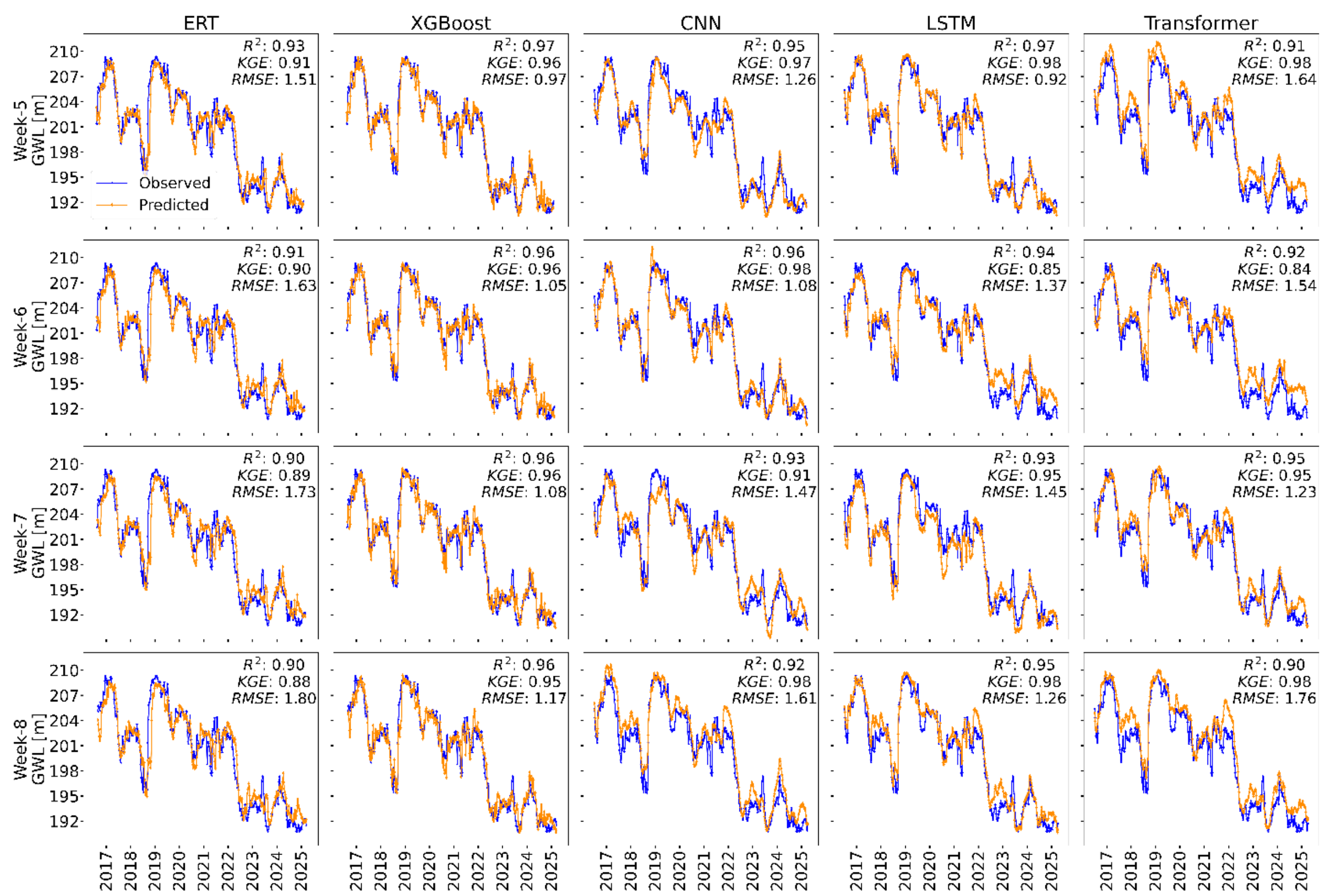

*Figure S11: Medium-term (5-8~week) predictions of groundwater levels at the J17 well. Observed (blue) and predicted (orange) groundwater levels are shown for all models. RMSE is expressed in m.*

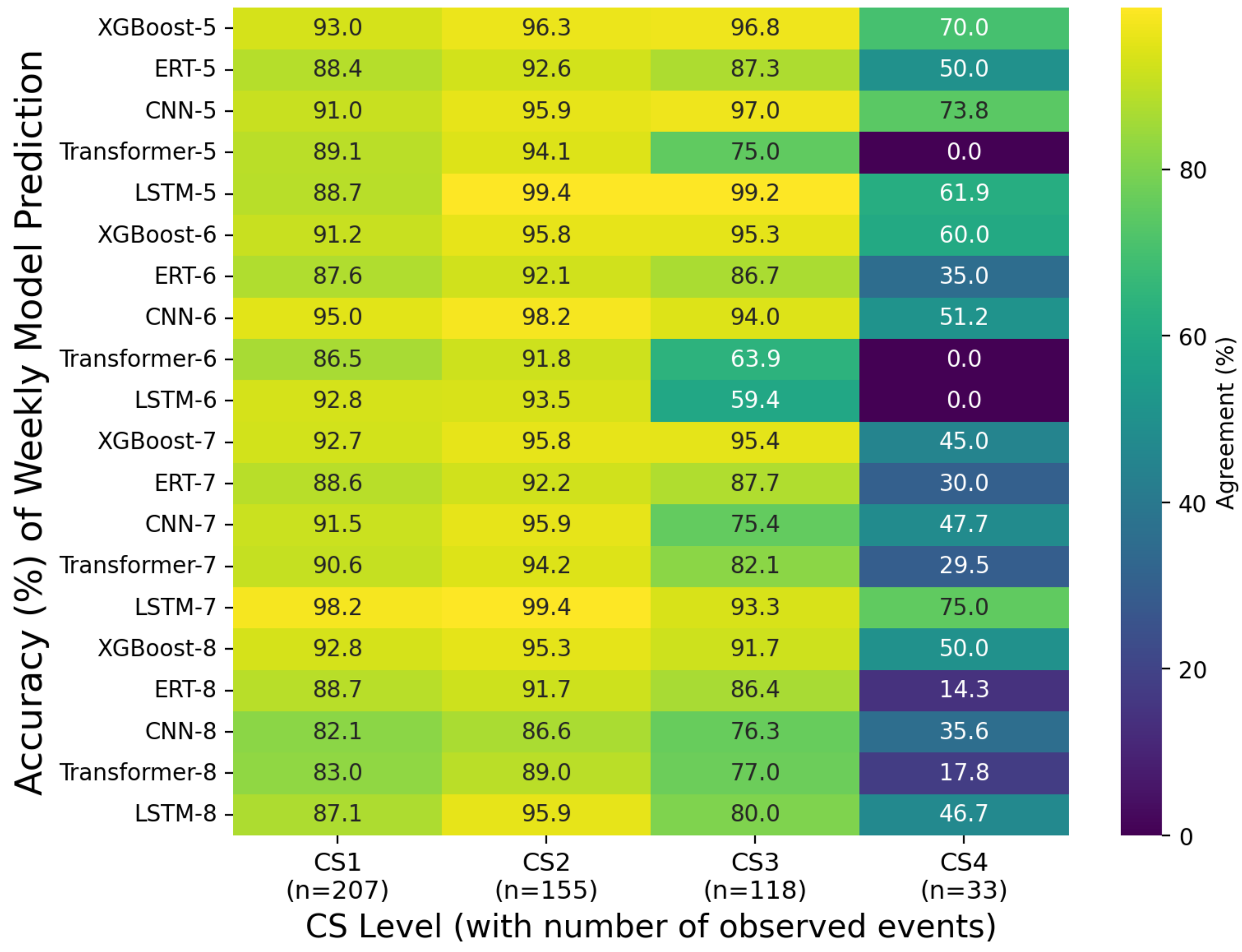

*Figure S12: Medium-term (5-8~weeks) groundwater level prediction performance of the AI models with respect to Critical Stage (CS) levels for the J17 well. Agreement indicates the percentage of times that spring flow values below a given CS threshold in the test dataset were correctly predicted.*

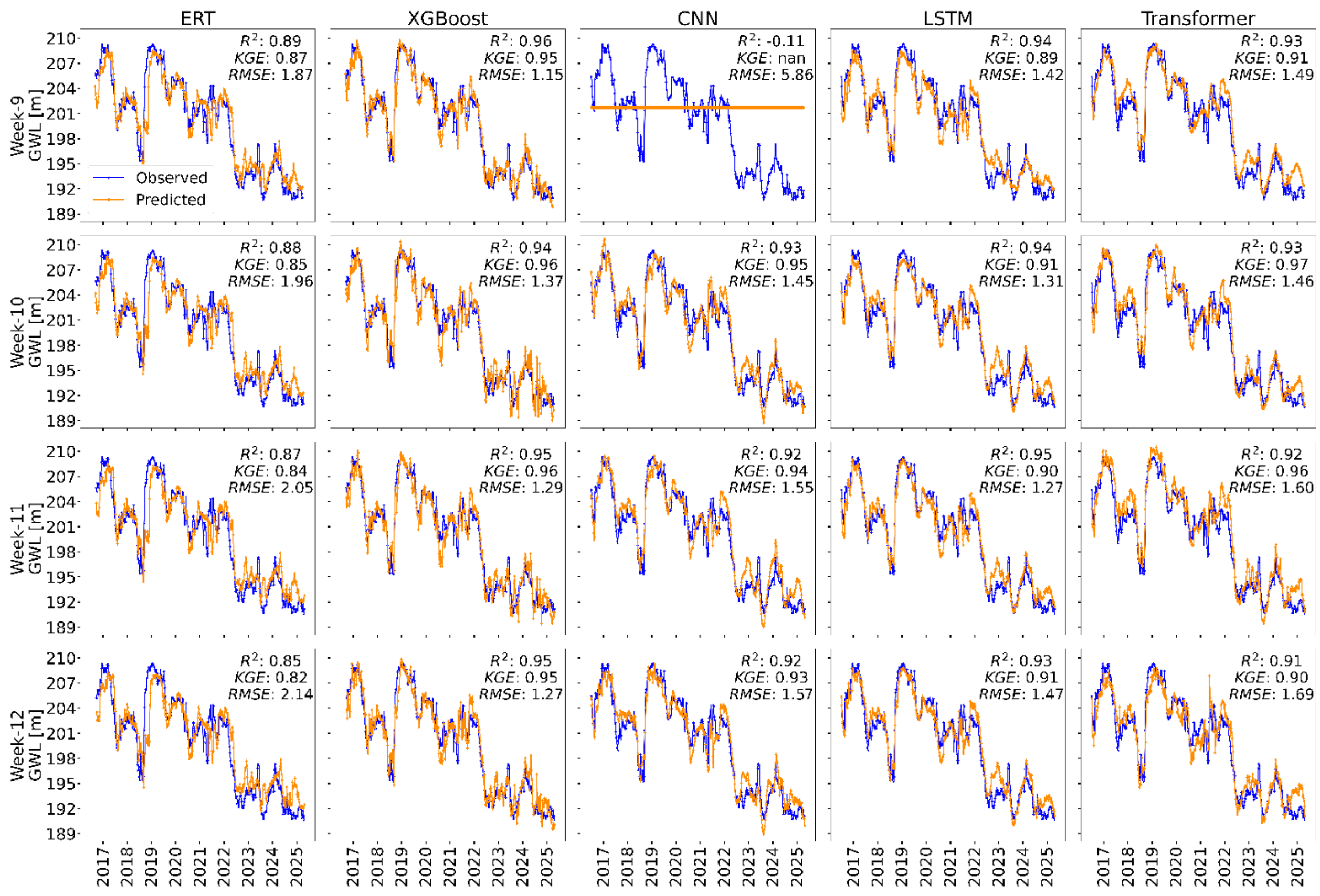

*Figure S13: Long-term (9--12 week) predictions of groundwater levels at the J17 well. Observed (blue) and predicted (orange) groundwater levels are shown for all models. RMSE is expressed in m.*

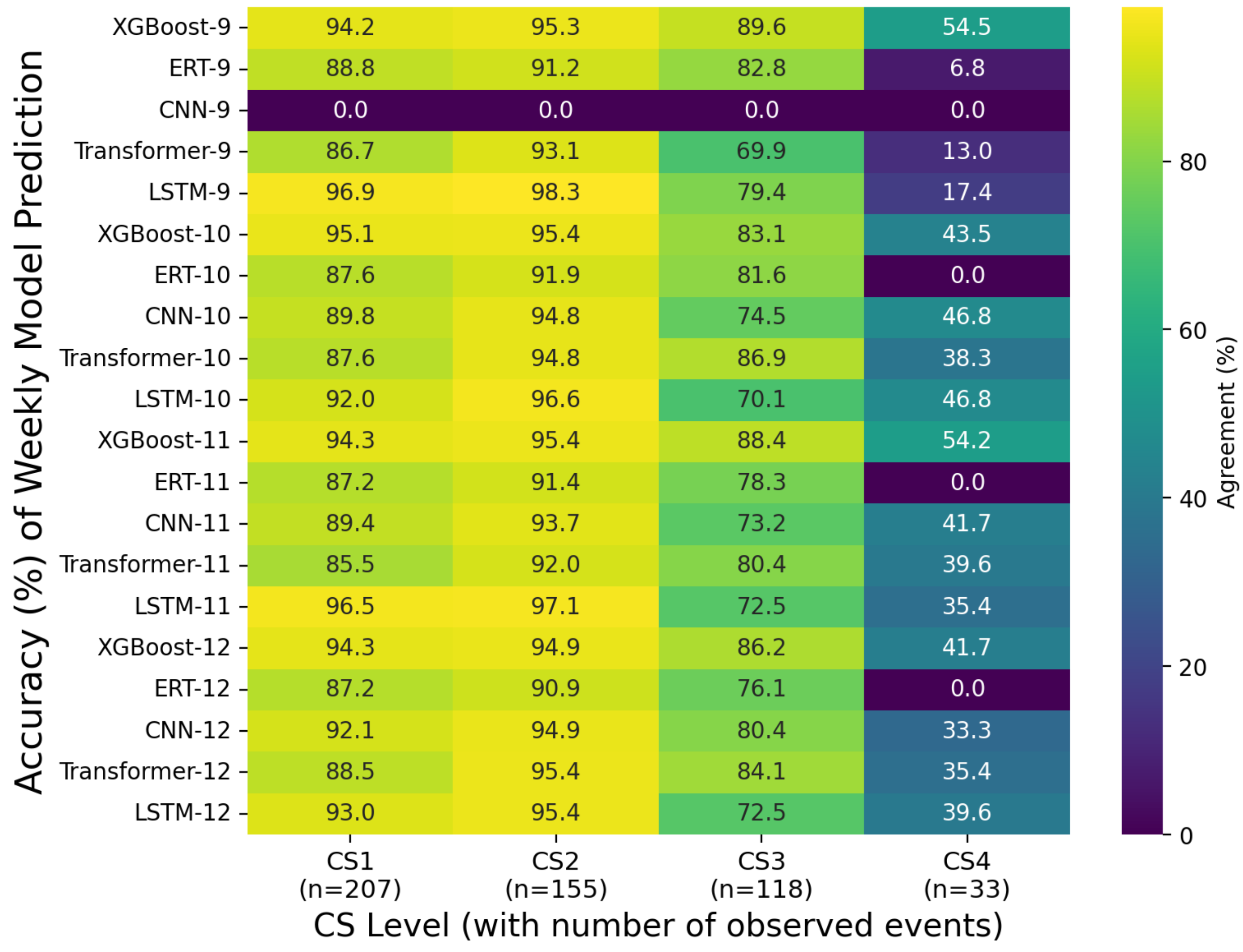

*Figure S14: Long-term (9-12 weeks) groundwater level prediction performance of the AI models with respect to Critical Stage (CS) levels for the J17 well. Agreement indicates the percentage of times that spring flow values below a given CS threshold in the test dataset were correctly predicted.*

Table S12: Hyperparameter grids for ETMs (XGBoost and ERT) used across lags 1–12.

| ***Hyperparameter*** | ***XGBoost*** | ***ERT*** |
|---|---|---|
| ***n_estimators*** | *20–5000 (5 values, equal interval)* | *20–10000 (5 values, equal interval)* |
| ***max_depth*** | *{2, 3, 4}* | *{4, 5, 6, 7}* |
| ***Learning rate*** | *0.01–0.33 (5 values, equal interval)* | — |

Table S13: Hyperparameter grids for DLMs (LSTM_seq, CNN_seq, Transformer_seq) used across lags 1–12.

| ***Hyperparameter*** | ***LSTM_seq*** | ***CNN_seq*** | ***Transformer_seq*** |
| --- | --- | --- | --- |
| ***Sequence length $t_s$*** | *{12, 26, 52}* | *{12, 26, 52}* | *{12, 26, 52}* |
| ***Learning rate (Adam)*** | *0.001–0.300 (10 values, equal interval)* | *0.001–0.300 (10 values, equal interval)* | *0.001–3.000 (10 values, equal interval)* |
| ***Batch size*** | *{32}* | *{32}* | *{32, 64}* |
| ***Model size*** | *LSTM units={64}; Dense={16}* | *filters={16, 32, 64}; kernel={3, 5}; Dense={16, 32, 64}* | *heads={2, 4}; key_dim={16, 32}; Dense={16, 32}* |
| ***Regularization*** | *Dropout (0.2, 0.2, 0.1)* | *GlobalAveragePooling1D* | *Dropout (0.2, 0.3), LayerNorm* |
| ***Optimizer / loss*** | *Adam / MSE* | *Adam / MSE* | *Adam / MSE* |
| ***Early stopping*** | *patience = 5* | *patience = 5* | *patience = 5* |

Table S14: Comal Springs Flow Forecasts and Accuracy Assessment, August 3–24, 2026

| ***Run Date*** | ***Lag (week)*** | ***Model*** | ***Target Week*** | ***Predicted (m³/s)*** | ***Observed (m³/s)*** | ***Error*** | ***% Error*** | ***Verified On*** |
|---|---|---|---|---|---|---|---|---|
| *2026-08-03* | *1* | ***XGB*** | *2026-08-09* | *5.8591* | *5.5501* | *+0.3090* | *5.57%* | *2026-08-11* |
| *2026-08-03* | *2* | ***XGB*** | *2026-08-16* | *5.4560* | *5.2386* | *+0.2174* | *4.15%* | *2026-08-17* |
| *2026-08-03* | *3* | ***XGB*** | *2026-08-23* | *5.2820* | *5.2103* | *+0.0717* | *1.38%* | *2026-08-24* |
| *2026-08-11* | *1* | ***XGB*** | *2026-08-16* | *5.1578* | *5.2386* | *-0.0808* | *-1.54%* | *2026-08-17* |
| *2026-08-11* | *2* | ***XGB*** | *2026-08-23* | *4.8491* | *5.2103* | *-0.3612* | *-6.93%* | *2026-08-24* |
| *2026-08-17* | *1* | ***XGB*** | *2026-08-23* | *4.8891* | *5.2103* | *-0.3212* | *-6.16%* | *2026-08-24* |

*Run Date = week the forecast was issued; Lag = forecast horizon in weeks; Model = XGB (XGBoost); Target Week = Monday of the forecast window; Error = Predicted − Observed; % Error = (Error / Observed) × 100. Green = positive error (over-prediction); Red = negative error (under-prediction).*

Table S15: J-17 Groundwater Level Forecasts and Accuracy Assessment, August 3–24, 2026

| ***Run Date*** | ***Lag (week)*** | ***Model*** | ***Target Week*** | ***Predicted (m)*** | ***Observed (m)*** | ***Error*** | ***% Error*** | ***Verified On*** |
|---|---|---|---|---|---|---|---|---|
| *2026-08-03* | *1* | ***XGB*** | *2026-08-09* | *196.9935* | *197.4708* | *-0.4773* | *-0.24%* | *2026-08-11* |
| *2026-08-03* | *2* | ***XGB*** | *2026-08-16* | *196.2339* | *196.8581* | *-0.6242* | *-0.32%* | *2026-08-17* |
| *2026-08-03* | *3* | ***XGB*** | *2026-08-23* | *196.0815* | *196.0260* | *+0.0555* | *0.03%* | *2026-08-24* |
| *2026-08-11* | *1* | ***XGB*** | *2026-08-16* | *196.5010* | *196.8581* | *-0.3571* | *-0.18%* | *2026-08-17* |
| *2026-08-11* | *2* | ***XGB*** | *2026-08-23* | *195.5917* | *196.0260* | *-0.4343* | *-0.22%* | *2026-08-24* |
| *2026-08-17* | *1* | ***XGB*** | *2026-08-23* | *195.7171* | *196.0260* | *-0.3089* | *-0.16%* | *2026-08-24* |
| *Run Date = week the forecast was issued; Lag = forecast horizon in weeks; Model = XGB (XGBoost); Target Week = Monday of the forecast window; Error = Predicted − Observed; % Error = (Error / Observed) × 100. Green = positive error; Red = negative error (under-prediction). All values in meters (m above mean sea level).* | | | | | | | | |